\documentclass[lettersize,journal]{IEEEtran}
\usepackage{subfigure}
\usepackage{amsmath,amsfonts}
\usepackage{algorithmic}
\usepackage{algorithm}
\usepackage{array}
\usepackage[caption=false,font=normalsize,labelfont=sf,textfont=sf]{subfig}
\usepackage{textcomp}
\usepackage{stfloats}
\usepackage{url}
\usepackage{verbatim}
\usepackage{graphicx}
\usepackage{cite}
\usepackage{multicol}
\usepackage{multirow}
\usepackage{makecell}
\usepackage{threeparttable}
\usepackage{amsfonts}
\usepackage{color,xcolor}
\usepackage{bm}

\begin{document}

\title{Very Lightweight Photo Retouching Network with Conditional Sequential Modulation}

\author{Yihao Liu$^\star$,
	Jingwen He$^\star$,
	Xiangyu Chen,
	Zhengwen Zhang,
	Hengyuan Zhao,\\
	Chao Dong,
	and Yu Qiao,~\IEEEmembership{Senior~Member,~IEEE}
	\IEEEcompsocitemizethanks{\IEEEcompsocthanksitem Y. Liu, J, He, X. Chen, Z. Zhang, H. Zhao, C. Dong and Y. Qiao are with Shenzhen Institute of Advanced Technology, Chinese Academy of Sciences, Shenzhen, 518055. E-mail: \{jw.he, hy.zhao1, chao.dong, yu.qiao\}@siat.ac.cn, \{chxy95, zhengwen.zhang02\}@gmail.com
		\IEEEcompsocthanksitem Y. Liu is also with the University of Chinese Academy of Sciences, Beijing, 100049. E-mail: liuyihao14@mails.ucas.ac.cn.
	}
	\thanks{$\star$ Y. Liu and J. He are co-first authors.}}



\maketitle

\begin{abstract}
Photo retouching aims at improving the aesthetic visual quality of images that suffer from photographic defects, especially for poor contrast, over/under exposure, and inharmonious saturation. In practice, photo retouching can be accomplished by a series of image processing operations. As most commonly-used retouching operations are pixel-independent, i.e., the manipulation on one pixel is uncorrelated with its neighboring pixels, we can take advantage of this property and design a specialized algorithm for efficient global photo retouching. We analyze these global operations and find that they can be mathematically formulated by a Multi-Layer Perceptron (MLP). Based on this observation, we propose an extremely lightweight framework -- Conditional Sequential Retouching Network (CSRNet). Benefiting from the utilization of $1\times1$ convolution, CSRNet only contains less than 37K trainable parameters, which are orders of magnitude smaller than existing learning-based methods. Experiments show that our method achieves state-of-the-art performance on the benchmark MIT-Adobe FiveK dataset quantitively and qualitatively. In addition to achieve global photo retouching, the proposed framework can be easily extended to learn local enhancement effects. The extended model, namely CSRNet-L, also achieves competitive results in various local enhancement tasks. Codes are available at \url{https://github.com/lyh-18/CSRNet}.
\end{abstract}


\section{Introduction}
\begin{figure*}[!htbp]
	\centering
	\includegraphics[width=1\linewidth,height=0.31\linewidth]{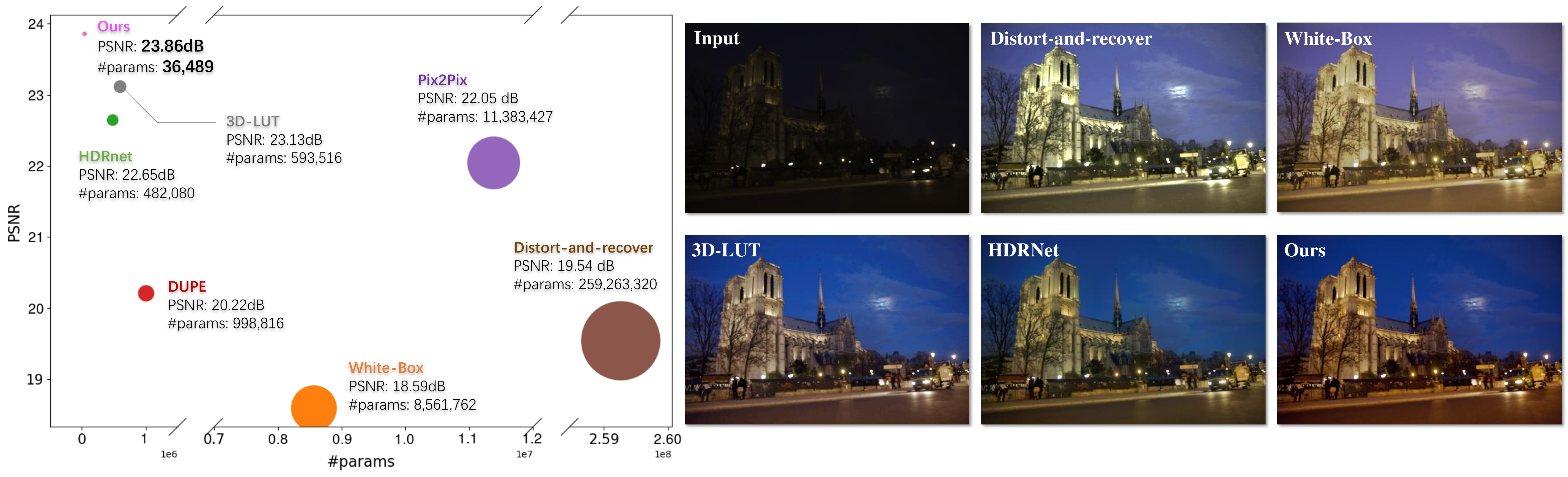}
	\caption{\textbf{Left: }Compared with existing state-of-the-art methods, our method achieves superior performance with much fewer parameters (1/13 of HDRNet \cite{hdrnet} and 1/250 of White-Box \cite{white-box}). The diameter of the circle represents the amount of trainable parameters. \textbf{Right: }Image retouching examples. Our method can generate more natural and visually pleasing retouched images than other methods. Best viewed in color.}
	\label{fig:1}
	
\end{figure*}

\IEEEPARstart{P}{hoto} retouching can significantly improve the visual quality of photographs through a sequence of image processing operations, such as brightness and contrast adjustments. Manual retouching requires specialized skills and training, thus is challenging for casual users. Even for professional retouchers, dealing with large collections requires tedious repetitive editing works. This presents the needs for automatic photo retouching method. It can be equipped in smart phones to help ordinary people get visual pleasing photos, or it can be built in photo editing software to provide an editing reference for experts.

The aim of photo retouching is to generate an aesthetically pleasing image from a low-quality input, which may suffer from under/over-exposure or unsaturated color tone. Recent learning-based methods tend to treat photo retouching as a special case of image enhancement or image-to-image translation. They use CNNs to learn either the transformation matrix \cite{hdrnet,3dlut} or an end-to-end mapping \cite{DSLR,WESPE,DPE} from input/output pairs. Generally, photo retouching needs to adjust the global color tones, while other image enhancement/translation tasks focus on local patterns or even change the image textures. Moreover, photo retouching is naturally a sequential processing, which can be decomposed into several simple operations. This property does not always hold for image enhancement and image-to-image translation problems. As most state-of-the-art algorithms \cite{hdrnet,Underexposed,DPE} are not specialized for photo retouching, they generally design complex network structures to deal with both global and local transformations, which largely restricts their performance and implementation efficiency. 

When investigating the retouching process, we find that most commonly-used retouching operations (e.g., white balancing, saturation controlling and color curve adjustment/tone-mapping) are global operations, without the change of local statistics. In real-world scenarios, global operations are essential for photo retouching, while local operations are optional. For example, the retouched images in the well-known benchmark MIT-Adobe FiveK dataset \cite{mit-adobe} contain only global tonal adjustment. The reinforcement learning (RL) based methods \cite{white-box, Distort} include only global operations in their action space. These motivate us to take advantage of global operations, and design an efficient algorithm especially for ``global'' photo retouching. After that, we can simply extend this method to achieve local retouching for other local operations. 

For preparation, we revisit several retouching operations adopted in RL based methods \cite{white-box, Distort}. Mathematically, these operations (e.g., white-balancing, saturation controlling, contrast adjustment) are \textbf{pixel-independent} and \textbf{location-independent}. In other words, the manipulation on one pixel is uncorrelated with neighboring pixels or pixels on specific positions. The input pixels can be mapped to the output pixels via \text{pixel-independent} mapping functions, without the need of local image features. We find that these pixel-wise functions can be approximated by a simple Multi-layer Perceptron (MLP). Different adjustment operations can share similar network structures but with different parameters. Then the input image can be sequentially processed by a set of MLPs to generate the final output.

Based on the above observation, we propose an extremely lightweight network - Conditional Sequential Retouching Network (CSRNet) - for fast global photo retouching. The key idea is to \textit{implicitly mimic the sequential processing procedure and model the editing operations in an end-to-end trainable network}. The framework consists of two modules - the base network and the condition network. The base network adopts a fully convolutional structure, while its unique feature is that all filters are of size $1\times1$, indicating that each pixel is processed independently. Therefore, the base network can be regarded as an MLP for individual pixels. To realize retouching operations, we modulate the intermediate features of the base network using the proposed Global Feature Modulation (GFM) layers, of which the parameters are controlled by the condition network. The condition network accepts the input image and generates a condition vector, which is then broadcasted to different layers of the base network for feature modulation. This procedure is just like a sequential editing process operated on different stages of the MLP. These two modules are jointly optimized with expert-adjusted image pairs. It is noteworthy that we name our method ``sequential modulation'' just for better understanding and illustration. In fact, our method does not explicitly model the step-wise retouching operations as RL-based methods. Instead, CSRNet implicitly models the whole process and generalizes well to unseen operations.

The proposed network is highly efficient for deployment in practical applications. It enjoys a very simple architecture, which contains only six plain convolutional layers in total, without any complex building blocks or even residual connections. Such a compact network could achieve state-of-the-art performance on MIT-Adobe FiveK dataset \cite{mit-adobe}, with less than 37K parameters -1/13 of HDRNet\cite{hdrnet} and 1/90 of DPE \cite{DPE} (see Figure \ref{fig:1} and Table \ref{table:main}). Note that even the first super-resolution CNN \cite{srcnn} with three layers already contains 56k parameters. We have also conducted extensive ablation studies on various settings, including different handcrafted global priors, network structures and feature modulation strategies. Further, in terms of feature modulation, we find it important to normalize the conditional inputs. We propose a novel and effective unit normalization (UN) to improve the training stability and enhance the performance. More explorations on the feature normalization and scaling techniques are also discussed.

While we have successfully achieved global photo retouching, the next step is to extend the proposed method to local enhancement operations. Note that CSRNet is very flexible in network design. We only need two simple modifications on the base network. The first one is to replace all $1\times1$ filters with larger ones, e.g., $3\times3$ filters. This could significantly enlarge the receptive field. The second one is to use the proposed Spatial Feature Modulation (SFM) to take the place of GFM. SFM is able to achieve spatial variant modulation, thus is more suitable for local operations. The modified framework, namely CSRNet-L, could handle both global and local retouching operations, and obtain comparable results against state-of-the-art methods in various local enhancement tasks.

A preliminary version of this work has been published in ECCV2020 \cite{csrnet}. The present work improves the initial version in significant ways. Firstly, we conduct more experiments and ablation studies to further improve the performance of CSRNet. Secondly, more analyses and explanations are added to the initial version. We make a demonstration experiment on simulating retouching operations, which clearly consolidates the theoretical analysis (Section \ref{sec:demonstration}). Thirdly, based on the proposed modulation network, we introduce a unit normalization (UN) operation on the generated condition vector. This operation can not only improve the training stability, but also bring further performance improvement. We also make an exploration on different feature normalization/scaling operations. Finally, we extend our method to handle local enhancement effects. With two simple modifications, the extended model CSRNet-L could achieve competitive performance with other methods in local enhancement tasks.

\section{Related Work}
Photo retouching and image enhancement has been studied for decades for its versatile applications in computer vision, image processing and aesthetic photograph editing \cite{hdrnet, Bilateralguided, retinex,dncnn,fdgan,hao2020low}. In this section, we briefly review the recent progress on image retouching and enhancement. Traditional algorithms have proposed various operations and filters to enhance the visual quality of images, such as histogram equalization, local Laplacian operator \cite{laplacian}, fast bilateral filtering \cite{Fastbilateral}, and color correction methods based on the gray-world \cite{Shades} or gray-edge \cite{Edge-based} assumption. Since Bychkovsky et al. \cite{mit-adobe} collected a large-scale dataset MIT-Adobe FiveK, which contains input and expert-retouched image pairs, a plenty of learning-based enhancing algorithms have been developed to continuously push the performance. Generally, these learning-based methods can be divided into three groups: physical-modeling-based methods, image-to-image translation methods and reinforcement learning methods. Physical-modeling-based methods attempt to estimate the intermediate parameters of the proposed physical models or assumptions for image enhancing. Based on the Retinex theory of color vision \cite{retinex}, several algorithms were developed for image exposure correction by estimating the reflectance and illumination with learnable models \cite{weightedvariational,cameraresponsemodel,correction,Underexposed,gao2017naturalness}. Zhang et al. \cite{zhang2020enhancing} adopted the Retinex theory and casted the enhancement problem as a constrained illumination estimation optimization. They formulated perceptually bidirectional similarity as constraints for underexposed photo enhancement. Zero-DCE \cite{guo2020zero} formulated light enhancement as a task of image-specific curve estimation with a deep network, which did not require any paired or unpaired data during training. Yan et al. \cite{automatic} adopted a multi-layer perceptron with a set of image descriptors to predict pixel-wise color transforms. Xu et al. \cite{xu2013generalized} established a generalized equalization model integrating contrast enhancement and white balancing into a unified framework of convex programming of image histogram.

 By postulating that the enhanced output image can be expressed as local pointwise transformations of the input image, Gharbi et al. \cite{hdrnet} combined bilateral grid \cite{bilateralgrid} and bilateral guided upsampling models \cite{Bilateralguided}, then constructed a CNN model to predict the affine transformation coefficients in bilateral space for real-time image enhancement.
Bianco et al. \cite{bianco2019learning} leveraged CNNs to learn parametric functions for color enhancement, which decoupled the inference of the parameters and the color transformation. SpliNet \cite{bianco2020personalized} estimated a global color transform for the enhancement of raw images. It predicted one set of control points from input raw image for each color channels and interpolated these control points with natural cubic splines. Recently, Zeng \textit{et al.} \cite{3dlut} proposed an image-adaptive 3-dimensional lookup tables (3D LUTs) to achieve fast and robust photo enhancement. It learns multiple basis 3D LUTs and a small convolutional neural network (CNN) simultaneously in an end-to-end manner.

Methods of the second group treat image enhancement as an image-to-image translation problem, which directly learn the end-to-end mapping between input and the enhanced image without modeling intermediate parameters. In practice, generative adversarial networks (GANs) have shown great potential for image transferring tasks \cite{pix2pix,CycleGAN,stargan,ranksrgan,esrgan}. Lee et al. \cite{lee2016automatic} introduced a method that can transfer the global color and tone statistics of the chosen exemplars to the input photo by selective style ranking technique from a large photo collection. Ignatov et al. explored to translate ordinary photos into DSLR-quality images by residual convolutional neural networks \cite{DSLR} and weakly supervised generative adversarial networks \cite{WESPE}. Yan et al. \cite{yan2014learning} formulated the color enhancement task as a learning-to-rank problem in which ordered pairs of images are used for training. Chen et al. \cite{DPE} utilized an improved two-way generative adversarial network (GAN) that can be trained in an unpair-learning manner. Zamir \textit{et al.} \cite{MIRNet} developed a multi-scale approach which maintains spatially precise high-resolution representations and receives strong contextual information from the low-resolution representations. Kim \textit{et al.}  \cite{glenet} combined a global enhancement network (GEN) and a local enhancement network (LEN) in one framework to achieve both paired and unpaired image enhancement. GEN performs the channel-wise intensity transform while LEN conducts spatial filtering to refine GEN results. Deng et al. \cite{deng2018aesthetic} proposed an aesthetic-driven image enhancement model with adversarial learning (EnhanceGAN), which requires weak supervision (binary labels on image aesthetic quality). Chai et al. \cite{chai2020supervised} extended the method of \cite{bianco2019learning} and pursued a GAN-based CNN that can be trained using either paired or unpaired images by determining the coefficients of a parametric color transformation. Ni et al. \cite{ni2020unpaired} developed a quality attention generative adversarial network (QAGAN), which was designed to learn domain-relevant quality attention directly from the low-quality and high-quality image domains. PieNet \cite{pienet} was the represented the users' preferences in latent vectors and then guide the network to achieve personalization.

Reinforcement learning is adopted for image retouching, which aims at explicitly simulating the step-wise retouching process. Hu et al. \cite{white-box} presented a White-Box photo post-processing framework that learns to make decisions based on the current state of the image. Park et al. \cite{Distort} casted the color enhancement problem into a Markov Decision Process (MDP) where each action is defined as a global color adjustment operation and selected by Deep Q-Network \cite{deepQ}. Yu et al. \cite{yu2018deepexposure} exploited deep reinforcement learning to learn multiple local exposure operations, in which an adversarial learning method is adopted to approximate the Aesthetic Evaluation (AE) function. In \cite{kosugi2020unpaired}, Satoshi and Toshihiko incorporated image editing software (such as Adobe Photoshop) into a GAN-based reinforcement learning framework, where the generator worked as the agent to select the software's parameters.

\section{Analysis of Retouching Operations}\label{sec:operations}
Photo retouching is accomplished by a series of image processing operations, such as the manipulation of brightness/contrast, the adjustment in each color channel, and the controlling of saturation/hue/tones. We mathematically find that these pixel-independent operations can be approximated or formulated by multi-layer perceptrons (MLPs). This observation motivates the design of our method. Below is our analysis on some representative retouching operations. The proposed framework is depicted in Section \ref{sec:method}.

\textbf{Global brightness change.} Given an input image $I$, the global brightness is described as the average value of its luminance map: $I_Y=0.299*I_R+0.587*I_G+0.114*I_B$, where $I_R$, $I_G$, $I_B$ represent the RGB channels, respectively. One simple way to adjust the brightness is to multiply a scalar for each pixel:
\begin{equation}\label{con:brightness}
I_{Y}^{'}(x,y)=\alpha I_Y(x,y), 
\end{equation}
where $I_{Y}^{'}(x,y)$ is the adjusted pixel value, $\alpha$ is a scalar, and $(x,y)$ indicates the pixel location in an $M \times N$ image. We can formulate the adjustment formula (\ref{con:brightness}) into the representation of an MLP:
\begin{equation}\label{mlp:brightness_mpl}
Y=f(\bm W^T \bm X+\bm b), 
\end{equation}
where $\bm X\in\mathbb{R}^{MN}$ is the vector flattened from the input image, $\bm W \in\mathbb{R}^{MN \times MN}$ and $\bm b\in\mathbb{R}^{MN}$ are weights and biases. $f(.)$ is the activation function. When $\bm W=diag\{\alpha,\alpha,\ldots,\alpha\}$, $\bm b=\vec{\bm 0}$ and $f$ is the identity mapping $f(\bm x)=\bm x$, the MLP (\ref{mlp:brightness_mpl}) is equivalent to the brightness adjustment formula (\ref{con:brightness}). 

\textbf{Contrast adjustment.} Contrast represents the difference in luminance or color maps. Among many definitions of contrast, we adopt a widely-used contrast adjustment formula:
\begin{equation} \label{con:contrast}
I^{'}(x,y)=\alpha I(x,y)+(1-\alpha)\overline{I},
\end{equation}
where $\overline{I}=\frac{1}{M \times N} \sum_{x=0}^{M-1}\sum_{y=0}^{N-1} I(x,y)$ and $\alpha$ is the adjustment coefficient. When $\alpha=1$, the image will remain the same. The above formula is applied on each channel of the image. We can construct a three-layer MLP that is equivalent to the contrast adjustment operation. For simplicity, \textit{the following derivation is for a single-channel image,} and it can be easily generalized to RGB images (refer to the derivation of white-balancing in the supplementary material). As in Figure \ref{fig:mlp}, the input layer has $M \times N$ units covering all pixels of the input image, the middle layer includes $M\times N+1$ hidden units and the last layer contains $M \times N$ output units. This can be formalized as:
\begin{equation}\label{mlp:contrast_mpl}
\begin{split}
Y=f_1(\bm W_1^T X+\bm b_1), Z=f_2(\bm W_2^T Y+\bm b_2),
\end{split}
\end{equation}
where $\bm X\in\mathbb{R}^{MN}$,  $\bm W_1\in\mathbb{R}^{MN \times (MN+1)}$,   $\bm W_2\in\mathbb{R}^{(MN+1) \times MN}$, $\bm b_1\in\mathbb{R}^{(MN+1)}$, $\bm b_2\in\mathbb{R}^{MN}$. Let $\bm A=diag\{\alpha,\alpha,\ldots,\alpha \}\in\mathbb{R}^{MN \times MN} $, $\bm B=\frac{1}{MN}\vec{\bm 1}\in\mathbb{R}^{MN}$, $\bm C=diag\{1,1,\ldots,1 \}\in\mathbb{R}^{MN \times MN}$, $\bm D=[(1-\alpha)\vec{\bm 1}]^T\in\mathbb{R}^{1 \times MN}$. When $\bm W_1=[A,B]\in\mathbb{R}^{MN \times (MN+1)}$, $\bm W_2=
\begin{bmatrix}
C \\
D
\end{bmatrix}\in\mathbb{R}^{(MN+1) \times MN} $,
$\bm b_1=\bm b_2=\vec{\bm 0}$ and $f_1(\bm x)=f_2(\bm x)=\bm x$,
the above MLP (\ref{mlp:contrast_mpl}) is equivalent to the contrast adjustment formula (\ref{con:contrast}).

Other operations, like white balancing, saturation controlling, color curve adjustment/tone-mapping, can also be regarded as MLPs. (Please refer to the supplementary materials.)

\textbf{Discussions.} So far, we have shown that most commonly-used retouching operations can be formulated as classic MLPs. These operations are pixel-independent and location-independent, i.e., \textit{the manipulation on one pixel is uncorrelated with its neighboring pixels or pixels on specific positions.} That is why we can use a diagonal matrix as the MLP weights. 
{Operations like brightness change, white-balancing, saturation controlling, tone-mapping, can be viewed as MLPs used \textit{{on a single pixel}}, which is similar with the MLPconv proposed in \cite{networkinnetwork}. The correlation between MLP and $1\times1$ convolutions has been revealed in MLPconv \cite{networkinnetwork} and SRCNN \cite{srcnn}. For contrast adjustment (Equ. \ref{con:contrast}), although it requires an additional image mean value in the operation, the mean value is not location-specific and is the same for any pixel to be manipulated. It actually can be considered as a constant. On the contrary, $3 \times 3$ convolution is not pixel-independent, since this operation involves neighboring pixels into the calculation.} Enlightened by this discovery, the base network in the proposed method is designed as a fully convolutional network with all the filter size of $1 \times 1$, which acts like an MLP worked on individual pixels and slides over the input image. Some operations, like contrast adjustment, may require global information that relates to all pixels in the image (e.g., image mean value). Such global information can be provided by the condition network in our method. {The condition network is proposed to extract global information from the input image to supplement the base network. The effectiveness of this design is further consolidated by simulation experiments in Section \ref{sec:demonstration}.}

\section{Methodology}\label{sec:method}
Our method aims at fast automatic image retouching with an extremely low parameter cost and computation consumption. Based on the analysis in Section \ref{sec:operations}, we propose a Conditional Sequential Retouching Network (CSRNet) which is specialized for efficient image retouching. It consists of a base network and a complementary condition network, which collaborate with each other to achieve image global tonal adjustment, as demonstrated in Section \ref{sec:experiment}. Then, we illustrate the intrinsic working mechanism of CSRNet in two comprehensive perspectives. We also describe how to achieve different retouching styles and to control the enhancing strength. Furthermore, our method can be easily extended to learning stylistic local effects.

\begin{figure}[htbp]
	\centering
	\subfigure[An MLP on individual pixels.]{		
		\includegraphics[width=2.5in]{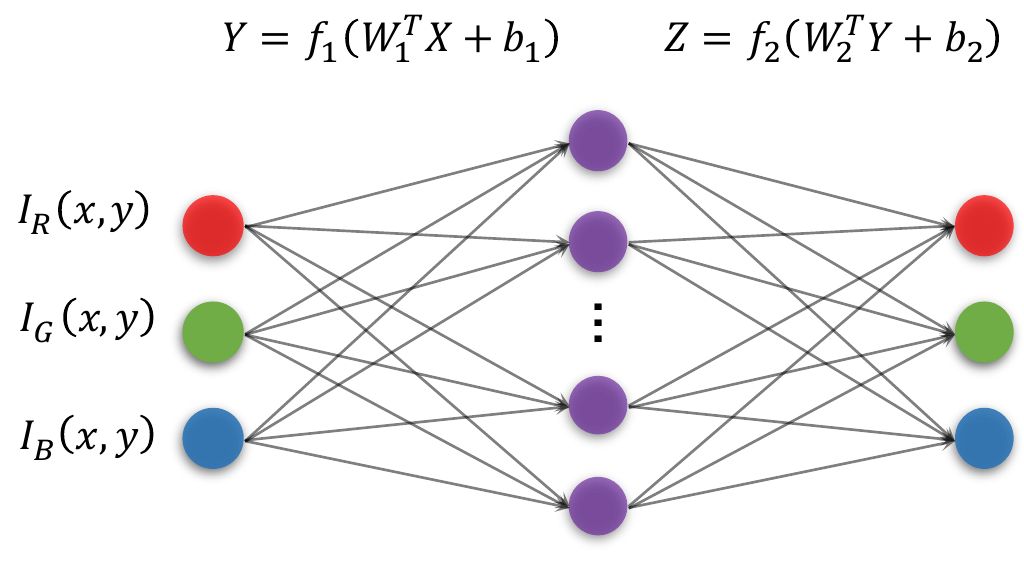}	
	}
	
	\subfigure[CSRNet. (k: kernel size; n: number of feature maps; s: stride.)]{
		\includegraphics[width=3in]{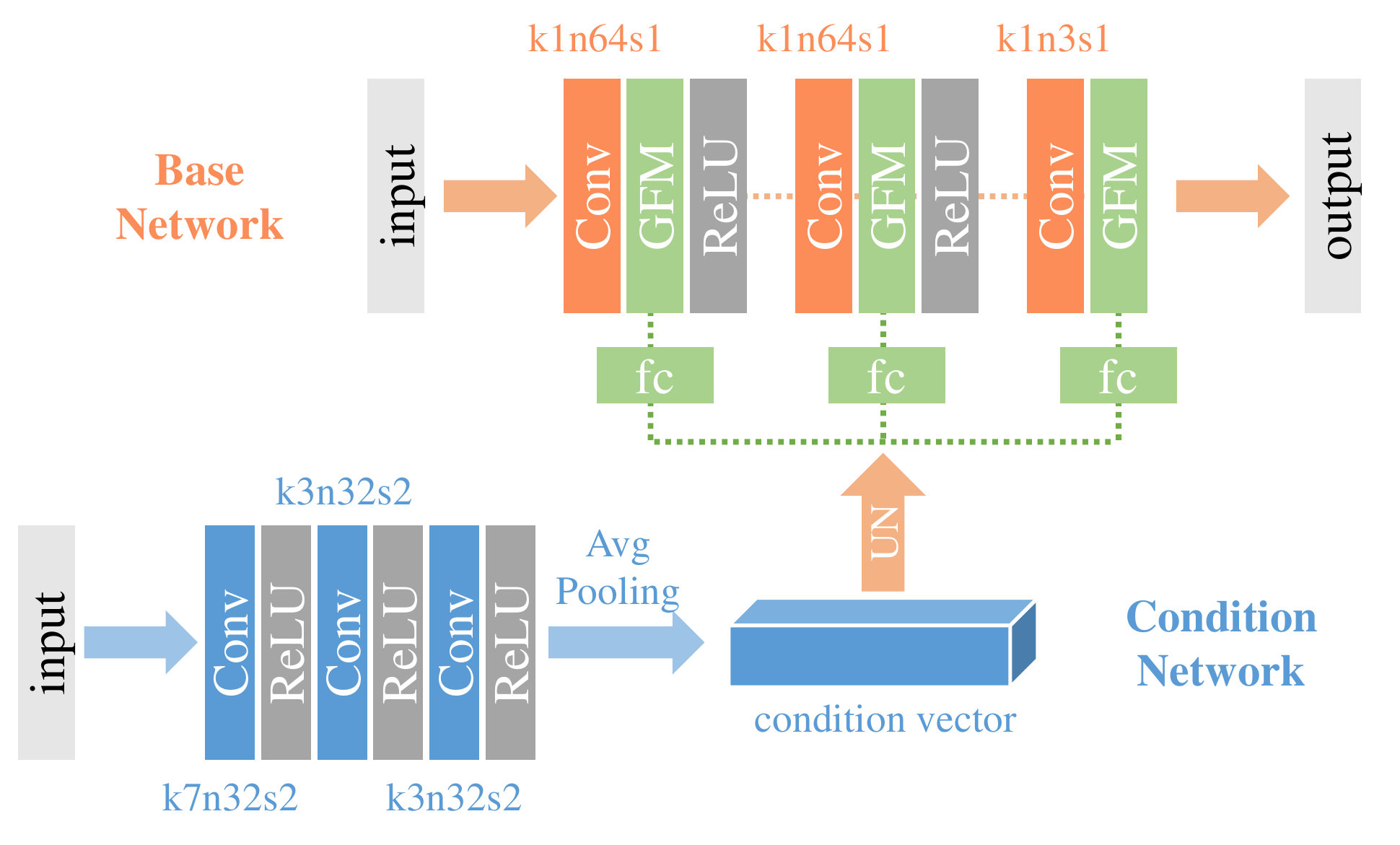}		
	}
	\caption{(a) Illustration for MLP on a single pixel. Pixel-independent operation can be viewed as an MLP used on individual pixels, such as brightness change, white-balancing, saturation controlling. (b) The proposed network consists of four key components -- base network, condition network, GFM and UN.
	}
	\label{fig:mlp}
\end{figure}

\subsection{Conditional Sequential Modulation Network}
The proposed conditional sequential retouching network (CSRNet) contains a base network and a condition network as shown in Figure \ref{fig:mlp}(b). The base network takes the low-quality image as input and generates the retouched image. The condition network estimates the global features from the input image, and afterwards influences the base network by conditional feature modulation operations.

\subsubsection{Network Structure} \label{sec:network}
The structures of the base network and the condition network are described as follows.

\textbf{Base network.} The base network adopts a fully convolutional structure with $N$ convolutional layers and $N-1$ ReLU activations. One unique trait of the base network is that all the filter size is $1\times1$, suggesting that each pixel in the input image is manipulated independently. Hence, the base network can be regarded as an MLP, which works on each pixel independently and slides over the input image, similar as MLPconv in \cite{networkinnetwork}. Based on the analysis in Section \ref{sec:operations}, theoretically, the base network has the capability of handling pixel-independent retouching operations. Moreover, since all the filters are of size $1\times1$, the network has dramatically few parameters.

\textbf{Condition network.} The global information/priors are indispensable for image retouching. For example, the contrast adjustment requires the average luminance of the image. To allow the base network to incorporate global features, a condition network is proposed to collaborate with the base network. The condition network is like an encoder that contains three blocks, in which a series of convolutional, ReLU and downsamping layers are included. The output of the condition network is a condition vector, which will be broadcasted into the base network using global feature modulation (GFM). Network details are depicted in Section \ref{sec:method} and Figure \ref{fig:mlp}(b).\\

\subsubsection{ Global Feature Modulation }
To enable the network to have the ability of handling operations that require global features, we modulate the intermediate features of the base network by element-wise multiplication and addition operations. The proposed conditional feature modulation can be formulated as follows:
\begin{equation}
GFM(\bm x) = \bm \gamma * \bm x + \bm \beta,
\end{equation}
where $*$ denotes the channel-wise broadcasting multiplication, $\bm x \in \mathbb{R}^{C \times H \times W}$ is the feature map to be modulated in the base network, and $\bm \gamma$, $\bm \beta \in \mathbb{R}^{C \times 1 \times 1}$ are affine parameters that are estimated from the outputs of the condition network. Since $\bm \gamma$ and $\bm \beta$ are vectors containing global statistics and the modulation is global-wise for each feature map in the base network, we call this operation Global Feature Modulation (GFM). Due to the effectiveness of GFM, the base network and condition network can collaborate well to achieve image global tonal adjustment with only few parameters.

\subsubsection{Unit Normalization}\label{sec:normalize}
To improve the stability and robustness, we propose a unit normalization (UN) operation on the condition vector to restrict its numerical range.
\begin{equation}
UN(\bm x_i) = \frac{\bm x_i}{\sqrt{\frac{1}{N}\sum_{k=1}^{N} \bm x_k^2}} = \sqrt{N}\frac{\bm x_i}{\left \| \bm x \right \|_2},
\end{equation}
where $x_i$ is an $N$-dimensional input vector. The unit normalization is similar with weight norm \cite{weightnorm}, but it acts on the feature vectors instead of the convolutional parameters. UN can be regarded as a kind of vector unitization with a scaling coefficient related to the vector dimension. Hence, this operation makes it focus more on the direction of the vector rather than the absolute value. After unit normalization, all the generated condition vectors fall on an $N$-sphere (hypersphere) with radius $\sqrt{N}$. In terms of feature modulation, we find UN operation plays an important role in the retouching performance. More explorations on the normalization/scaling strategies are demonstrated in Section \ref{sec:explore_normalization}.

{\textbf{Comparison with StyleGAN and SFT-GAN.} Notably, the proposed condition network plays different roles and have different motivations with StyleGAN \cite{styleGAN} and SFT-GAN \cite{sftnet}. Specifically, the $1 \times 1$ base network is devised for efficiency and the supplementary condition network is proposed to extract the global image information to tackle retouching operations that require global statistics (e.g., contrast adjustment). In StyleGAN, the style code is mainly used to embed the image high-level attributes (e.g., pose, smile, identity), and to adjust the ``style'' of the image at each convolution layer. In addition, the latent code in StyleGAN is sampled and transformed from a normal distribution. In SFT-GAN \cite{sftnet}, segmentation maps are introduced to guide the SR network to produce more semantic-aware textures (e.g., plant, building). Its modulation parameters are transformed from the segmentation probability maps, while our condition vector is directly extracted from the input image. In summary, our condition network and the feature modulation strategy are conceptually different from  StyleGAN and SFT-GAN in motivation and implementation.}

\subsection{Extension to Learning Local Effects}\label{sec:local}
\begin{figure}[h]
	\centering
	\includegraphics[width=3.3in]{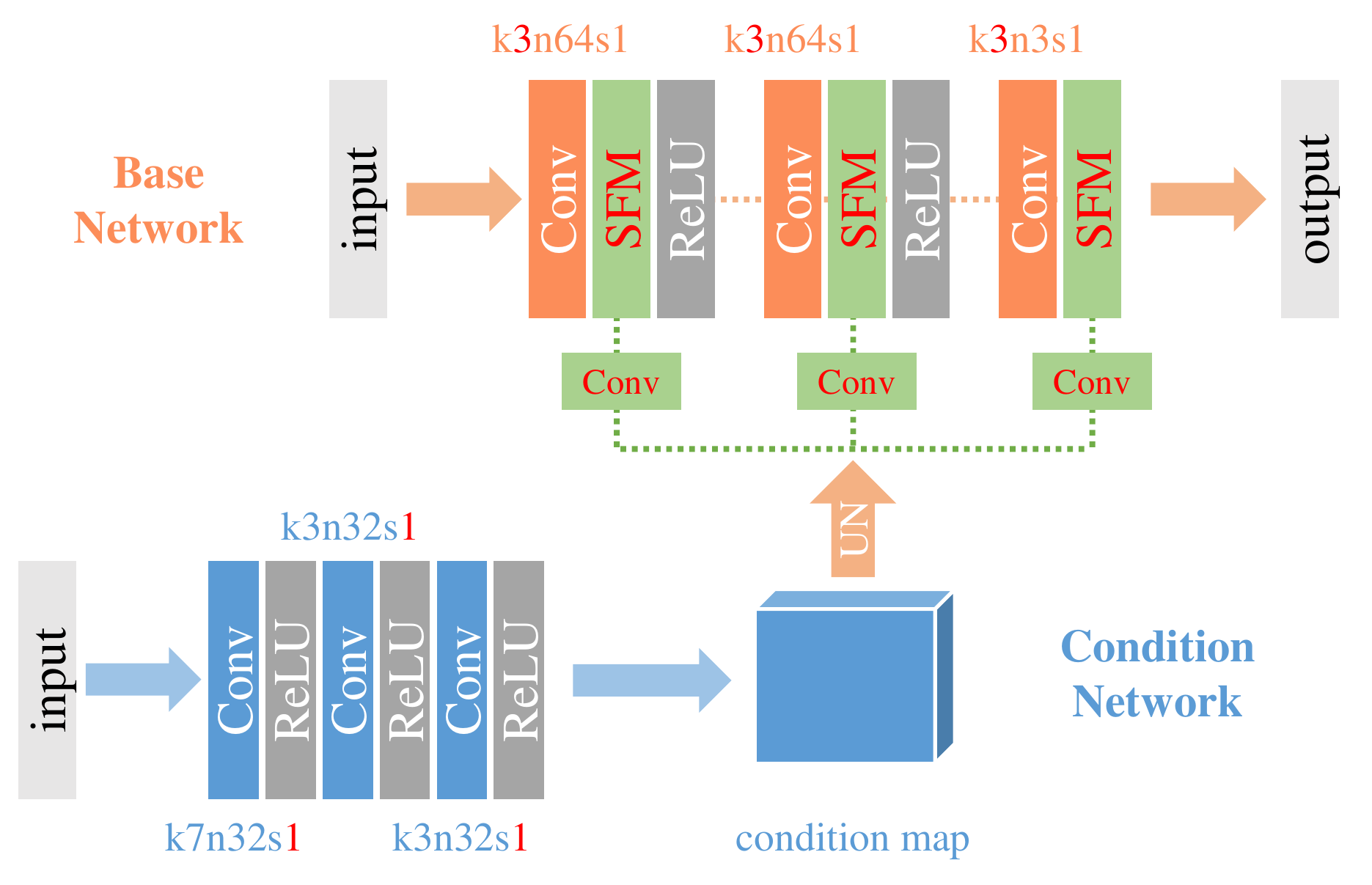}
	\caption{The framework of CSRNet-L. Comparing with CSRNet, there are three major modifications for achieving local adjustment. 1) Enlarge the filter size in the base network. 2) The condition network does not downsample the feature maps, thus will output a condition map rather than a vector. 3) Adopt spatial feature modulation strategy. The modified parts in the framework are highlighted in red color.}
	\label{fig:frame_csrnetl}
	
\end{figure}

Thanks to the utilization of $1 \times 1$ convolutions and the corresponding condition network, the proposed method only contains a few parameters and is efficient for global tonal adjustment. Besides, our method can also be extended to learn local enhancement through expanding the base network and adopting spatial feature modulation strategy. The extended network that could achieve local effects is denoted as CSRNet-L. The overall framework is depicted in Figure \ref{fig:frame_csrnetl}.

\subsubsection{Framework of CSRNet-L}
Local adjustments require complex spatially varying manipulations on local regions, such as adjusting local contrast, performing stylistic effects and stressing the foreground color tone. To realize such local adjustments, the model needs to be capable of learning to capture local patterns and perform location-aware operations. Although our method is proposed for global image adjustments, it can be easily extended to learn complicated local adjustments with two simple modifications, which are marked with red color in Figure \ref{fig:frame_csrnetl}. 

\textbf{Expanding the base network.} $1 \times 1$ convolution is not enough to handle local enhancement, since it manipulates pixels independently without considering neighborhood information. We increase the filter size of the base network to $3 \times 3$, so that it can learn more complex nonlinear mappings and deal with local patterns. 

\textbf{Spatial feature modulation.} Local effects require spatially variant operations on different regions. However, convolution operations are spatially invariant, which means that the same convolution filter weights are utilized on all positions. This property violates the goal of local enhancement that manipulates images according to pixel locations. Furthermore, once the network is trained, all the filter weights are then fixed and applied to all image samples. To this end, we extend the global feature modulation (GFM) to spatial feature modulation (SFM), which is first introduced in SFT-Net \cite{sftnet} for semantic super resolution. The formula of SFM is similar as GFM, which is shown as
\begin{equation}
SFM(x) = \bm \gamma \odot \bm x + \bm \beta,
\end{equation}
where $\odot$ denotes the element-wise multiplication. $\bm x \in \mathbb{R}^{C \times H \times W}$ is the feature maps to be modulated in the base network, and $\bm \gamma$, $\bm \beta \in \mathbb{R}^{C \times H \times W}$ are learned modulation parameters. Note that, in GFM, the modulation parameters are only $C$-dimensional vectors for global adjustment. When using SFM in the base network, the condition network will output a set of condition maps instead of condition vectors, as shown in Figure \ref{fig:frame_csrnetl}. Hence, SFM can help the network perform spatially variant and location-specific manipulations.

\section{Learning Global Adjustments} \label{sec:experiment}
\subsection{Experimental Setup}
\textbf{Dataset and Metrics.} MIT-Adobe FiveK \cite{mit-adobe} is a commonly-used photo retouching dataset with 5, 000 RAW images and corresponding retouched versions produced by five experts (A/B/C/D/E). We follow the previous methods \cite{white-box,DPE,Underexposed,hdrnet} to use the retouched results of expert C as the ground truth (GT). We adopt the same pre-processing procedure as \cite{white-box} \footnote{https://github.com/yuanming-hu/exposure/wiki/Preparing-data-for-the-MIT-Adobe-FiveK-Dataset-with-Lightroom} and all the images are resized to 500px on the long edge. We randomly select 500 images for testing and the remaining 4,500 images for training. We use PSNR, SSIM and the Mean L2 error in {CIE L*a*b*} space \footnote{CIE L*a*b* (CIELAB) is a color space specified by the International Commission on Illumination. It describes all the colors visible to the human eye and was created to serve as a device-independent model to be used as a reference.} {(also known as $\Delta E$)} to evaluate the performance. \\
\textbf{Implementation Details.}
The base network contains 3 convolutional layers with channel size 64 and kernel size $1\times1$. The condition network also contains three convolutional layers with channel size 32. The kernel size of the first convolutional layer is set to $7\times7$ to increase the receptive field, while others are $3\times3$. Each convolutional layer downsamples features to half size with a stride of 2. We use a global average pooling layer at the end of the condition network to obtain a 32-dimensional condition vector. Then the condition vector will be transformed by fully connected layers to generate the parameters of channel-wise scaling and shifting operations. In total, there are 6 fully connected layers for 3 scaling operations and 3 shifting operations. During training, the mini-batch size is set to 1. L1 loss is adopted as the loss function. The learning rate is initialized as $10^{-4}$ and is decayed by a factor of 2 every $1.5 \times 10^{5}$ iterations. All experiments run $6\times10^{5}$ iterations. We use PyTorch framework and train all models on GTX 2080Ti GPUs. It takes only 5 hours to train the model.

\subsection{Simulating Retouching Operations} \label{sec:demonstration}
To support the analysis in Section \ref{sec:operations} and Section \ref{sec:method}, we use the proposed network to simulate the procedures of several retouching operations, including global brightness change, tone-mapping and contrast adjustment. Specifically, we adopt images retouched by expert C as inputs and apply retouching operations with specified adjustment parameters on the inputs as supervision labels. Then we utilize the base network and the proposed CSRNet to learn such mappings. 

Theoretically, the base network can perfectly handle operations like global brightness change and tone-mapping, because these pixel-independent operations are equivalent to MLPs used on individual pixels. For contrast adjustment, only the base network should not be enough, since it cannot extract global information like image mean value.

\setlength{\tabcolsep}{5pt}
\begin{table}
	\begin{center}
		\caption{Demonstration experiment on simulating retouching operations. Our method can successfully handle commonly-used retouching operations, which is consistent with the theoretical analysis. The results in ``contrast'' adjustment also show the significance of adopting the condition network.}
		\label{table:demonstration}
		\renewcommand{\arraystretch}{1.4}
		\begin{threeparttable}
			\begin{tabular}{ccccc}
				\hline\noalign{\smallskip}
				Operations & \makecell[c]{Original \\ (Input-GT)} & \makecell[c]{Base\\Netwok} & \makecell[c]{Condition\\Netwok} & PSNR \\
				\noalign{\smallskip}
				\hline
				\noalign{\smallskip}
				\multirow{2}*{\makecell[c]{brightness \\($\alpha=1.5$) }} & \checkmark &  $\times$ & $\times$ & 14.7413 \\
				& $\times$ &\checkmark & $\times$ & \textbf{69.7061}  \\
				\hline
				\multirow{2}*{\makecell[c]{brightness \\($\alpha=0.5$) }} & \checkmark &  $\times$ & $\times$ & 12.8460  \\
				& $\times$ &\checkmark & $\times$ & \textbf{69.0525}  \\
				\hline
				\multirow{2}*{\makecell[c]{tone-mapping\tnote{*} \\($L=4$)\\ }} & \checkmark &  $\times$ & $\times$ & 21.7580  \\
				& $\times$ &\checkmark & $\times$ & \textbf{56.1175}  \\
				\hline
				\multirow{3}*{\makecell[c]{contrast \\($\alpha=1.5$) }} & \checkmark &  $\times$ & $\times$ & 21.3584  \\
				& $\times$ &\checkmark & $\times$ & 28.6734  \\
				& $\times$ &\checkmark & \checkmark & \textbf{60.5206}\\
				\hline	
			\end{tabular}
			\begin{tablenotes}
				\item[*] The parameters for tone-mapping are set to $t_i=[3/8,2/8,1/8,2/8]$.
			\end{tablenotes}
		\end{threeparttable}
	\end{center}
\end{table}

The results are shown in Table \ref{table:demonstration}. As expected, the base network can successfully deal with the pixel-independent operations \footnote{Images are basically the same when PSNR $\textgreater$ 50dB.}. Nevertheless, we observe that a sole base network is unable to handle contrast adjustment, which requires global information. We can solve this problem by introducing the condition network. As we can see, the PSNR rises from 28dB to 60dB, demonstrating the effectiveness of introducing a condition network for providing more supportive global information. This simulation experiment consolidates the theoretical analysis and the practical design of the proposed framework.

\begin{figure*}[htbp]
	\centering
	\begin{minipage}[t]{0.18\textwidth}
		\centering
		\includegraphics[width=3.25cm,height=2.7cm]{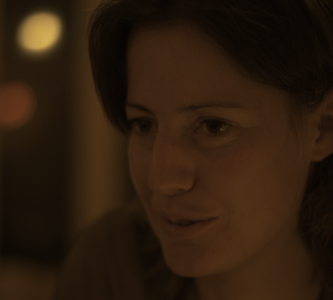}\\
		\scriptsize input
	\end{minipage}
	\begin{minipage}[t]{0.18\textwidth}
		\centering
		\includegraphics[width=3.25cm,height=2.7cm]{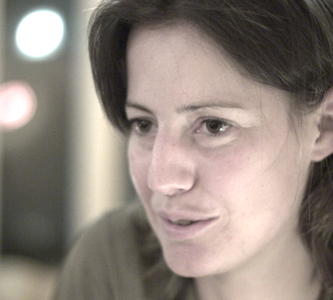}\\
		\scriptsize Distort-and-recover
	\end{minipage}
	\begin{minipage}[t]{0.18\textwidth}
		\centering
		\includegraphics[width=3.25cm,height=2.7cm]{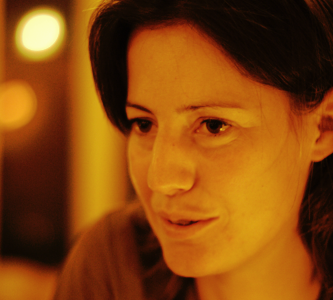}\\
		\scriptsize White-box    	
	\end{minipage}
	\begin{minipage}[t]{0.18\textwidth}
		\centering
		\includegraphics[width=3.25cm,height=2.7cm]{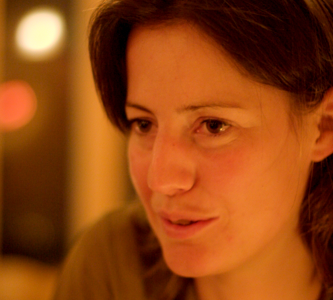}\\
		\scriptsize DPE
	\end{minipage}
	\begin{minipage}[t]{0.18\textwidth}
		\centering
		\includegraphics[width=3.25cm,height=2.7cm]{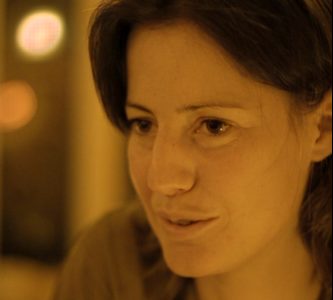}\\
		\scriptsize MIRNet
	\end{minipage}\\
	
	\begin{minipage}[t]{0.18\textwidth}
		\centering
		\includegraphics[width=3.25cm,height=2.7cm]{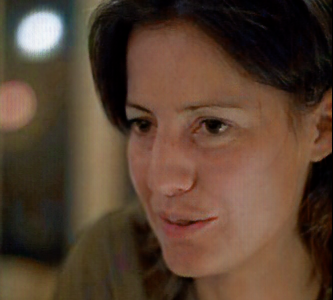}\\
		\scriptsize Pix2Pix
	\end{minipage}
	\begin{minipage}[t]{0.18\textwidth}
		\centering
		\includegraphics[width=3.25cm,height=2.7cm]{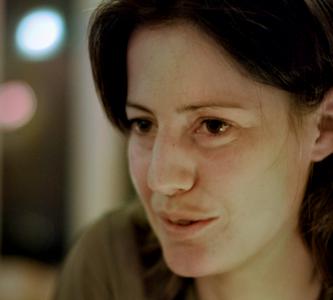}\\
		\scriptsize HDRNet
	\end{minipage}
	\begin{minipage}[t]{0.18\textwidth}
		\centering
		\includegraphics[width=3.25cm,height=2.7cm]{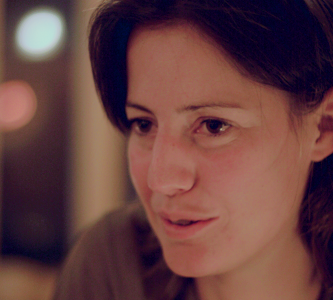}\\
		\scriptsize 3D-LUT
	\end{minipage}
	\begin{minipage}[t]{0.18\textwidth}
		\centering
		\includegraphics[width=3.25cm,height=2.7cm]{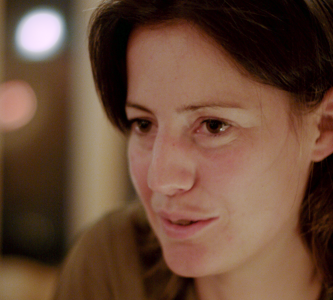}\\
		\scriptsize CSRNet (ours)
	\end{minipage}
	\begin{minipage}[t]{0.18\textwidth}
		\centering
		\includegraphics[width=3.25cm,height=2.7cm]{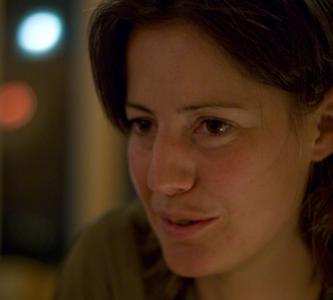}\\
		\scriptsize GT
	\end{minipage}\\
	
	\begin{minipage}[t]{0.18\textwidth}
		\centering
		\includegraphics[width=3.25cm]{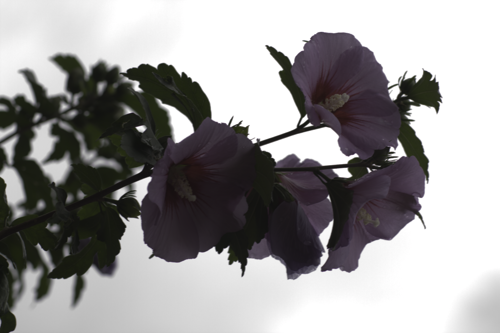}\\
		\scriptsize{input}
	\end{minipage}
	\begin{minipage}[t]{0.18\textwidth}
		\centering
		\includegraphics[width=3.25cm]{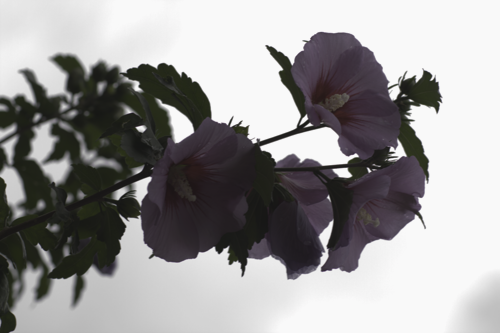}\\
		\scriptsize{Distort-and-recover}
	\end{minipage}
	\begin{minipage}[t]{0.18\textwidth}
		\centering
		\includegraphics[width=3.25cm]{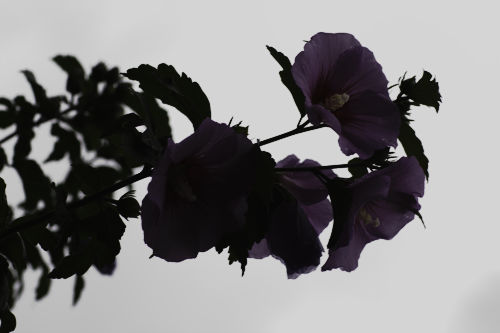}\\
		\scriptsize{White-box}
	\end{minipage}
	\begin{minipage}[t]{0.18\textwidth}
		\centering
		\includegraphics[width=3.25cm]{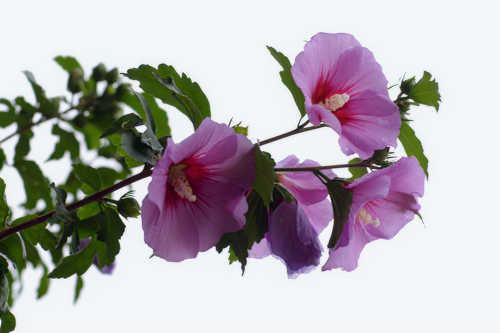}\\
		\scriptsize{DPE}
	\end{minipage}
	\begin{minipage}[t]{0.18\textwidth}
		\centering
		\includegraphics[width=3.25cm]{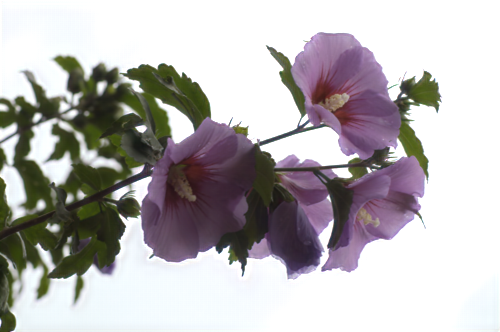}\\
		\scriptsize{MIRNet}
	\end{minipage}\\
	
	\begin{minipage}[t]{0.18\textwidth}
		\centering
		\includegraphics[width=3.25cm]{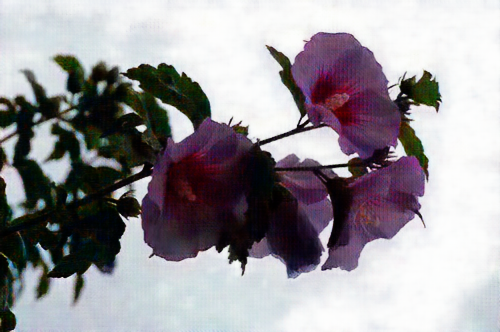}\\
		\scriptsize{Pix2Pix}
	\end{minipage}
	\begin{minipage}[t]{0.18\textwidth}
		\centering
		\includegraphics[width=3.25cm]{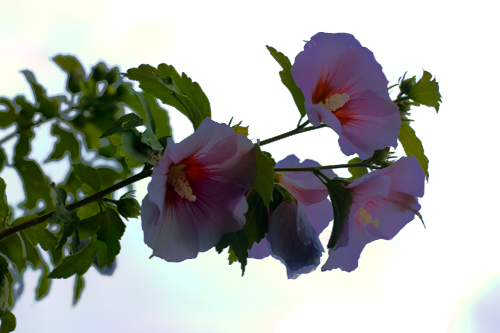}\\
		\scriptsize{HDRNet}
	\end{minipage}
	\begin{minipage}[t]{0.18\textwidth}
		\centering
		\includegraphics[width=3.25cm]{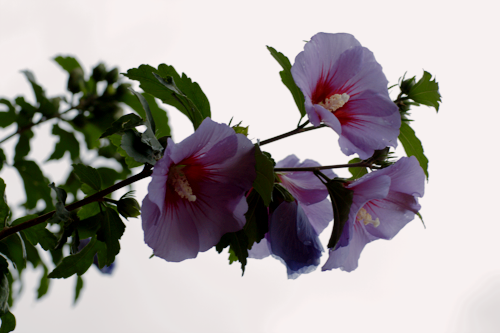}\\
		\scriptsize{3D-LUT}
	\end{minipage}
	\begin{minipage}[t]{0.18\textwidth}
		\centering
		\includegraphics[width=3.25cm]{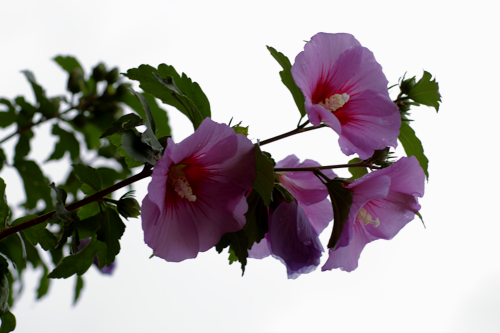}\\
		\scriptsize{CSRNet (ours)}
	\end{minipage}
	\begin{minipage}[t]{0.18\textwidth}
		\centering
		\includegraphics[width=3.25cm]{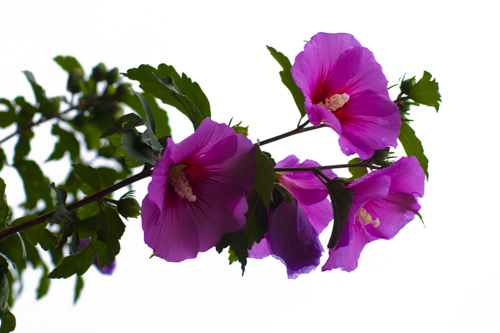}\\
		\scriptsize{GT}
	\end{minipage} \\
	
	\begin{minipage}[t]{0.18\textwidth}
		\centering
		\includegraphics[width=3.25cm]{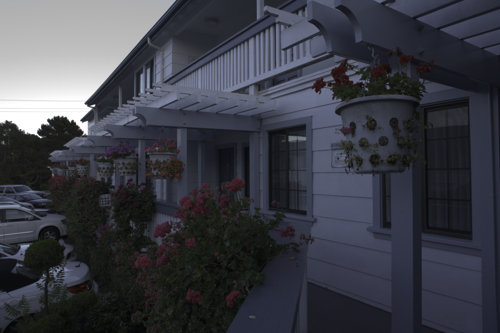}\\
		\scriptsize{input}
	\end{minipage}
	\begin{minipage}[t]{0.18\textwidth}
		\centering
		\includegraphics[width=3.25cm]{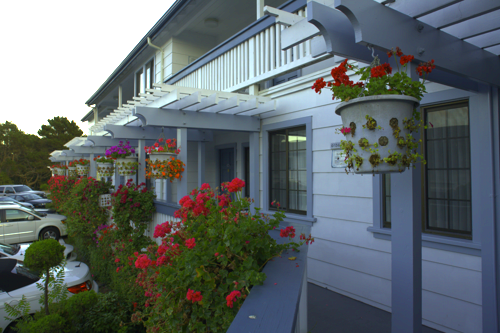}\\
		\scriptsize{Distort-and-recover}
	\end{minipage}
	\begin{minipage}[t]{0.18\textwidth}
		\centering
		\includegraphics[width=3.25cm]{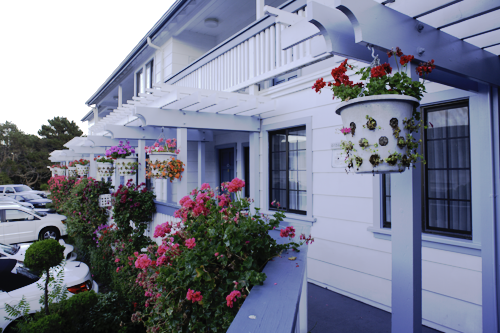}\\
		\scriptsize{White-box}
	\end{minipage}
	\begin{minipage}[t]{0.18\textwidth}
		\centering
		\includegraphics[width=3.25cm]{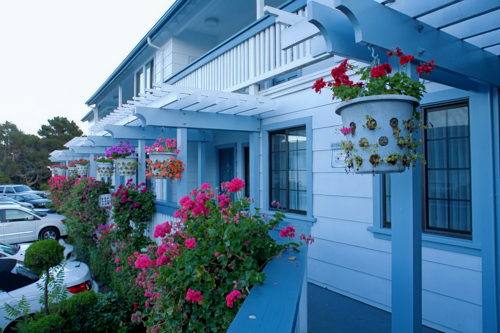}\\
		\scriptsize{DPE}
	\end{minipage}
	\begin{minipage}[t]{0.18\textwidth}
		\centering
		\includegraphics[width=3.25cm]{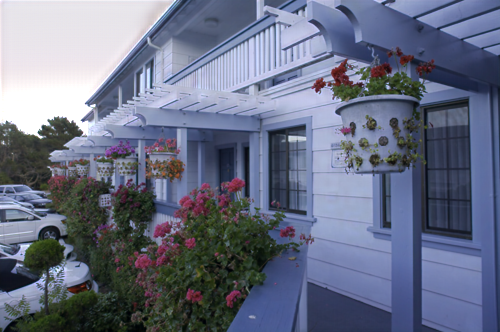}\\
		\scriptsize{MIRNet}
	\end{minipage}\\
	
	\begin{minipage}[t]{0.18\textwidth}
		\centering
		\includegraphics[width=3.25cm]{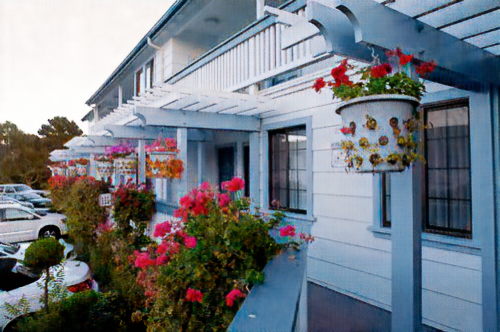}\\
		\scriptsize{Pix2Pix}
	\end{minipage}
	\begin{minipage}[t]{0.18\textwidth}
		\centering
		\includegraphics[width=3.25cm]{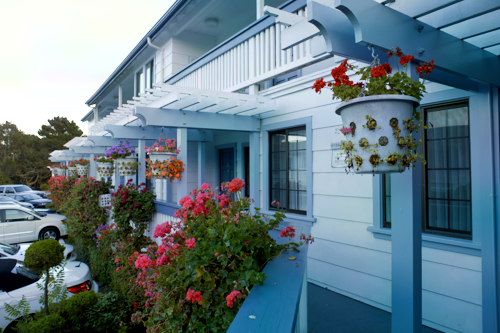}\\
		\scriptsize{HDRNet}
	\end{minipage}
	\begin{minipage}[t]{0.18\textwidth}
		\centering
		\includegraphics[width=3.25cm]{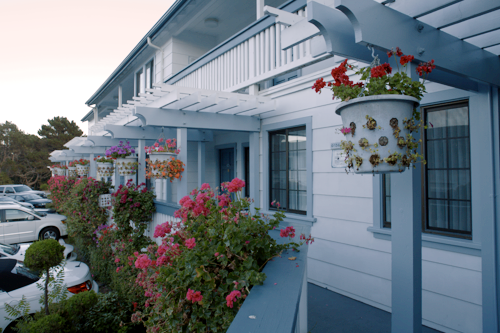}\\
		\scriptsize{3D-LUT}
	\end{minipage}
	\begin{minipage}[t]{0.18\textwidth}
		\centering
		\includegraphics[width=3.25cm]{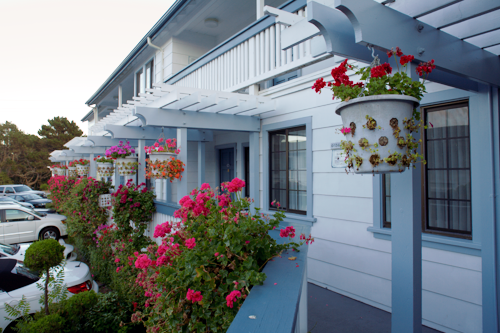}\\
		\scriptsize{CSRNet (ours)}
	\end{minipage}
	\begin{minipage}[t]{0.18\textwidth}
		\centering
		\includegraphics[width=3.25cm]{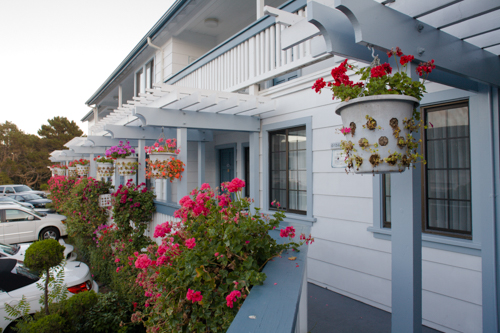}\\
		\scriptsize{GT}
	\end{minipage} \\
	
	\begin{minipage}[t]{0.18\textwidth}
		\centering
		\includegraphics[width=3.25cm]{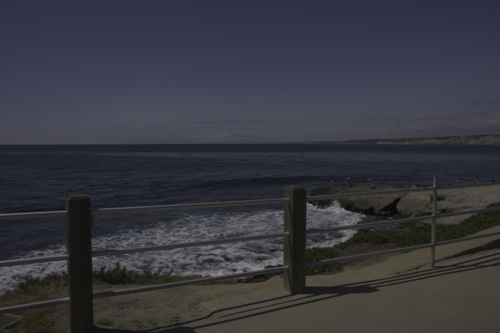}\\
		\scriptsize{input}
	\end{minipage}
	\begin{minipage}[t]{0.18\textwidth}
		\centering
		\includegraphics[width=3.25cm]{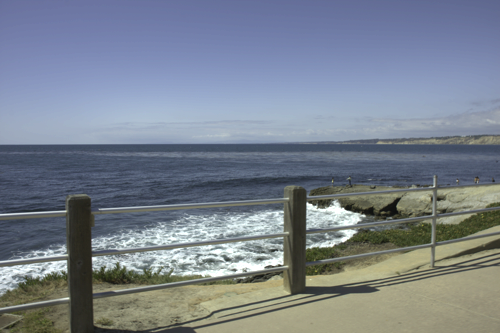}\\
		\scriptsize{Distort-and-recover}
	\end{minipage}
	\begin{minipage}[t]{0.18\textwidth}
		\centering
		\includegraphics[width=3.25cm]{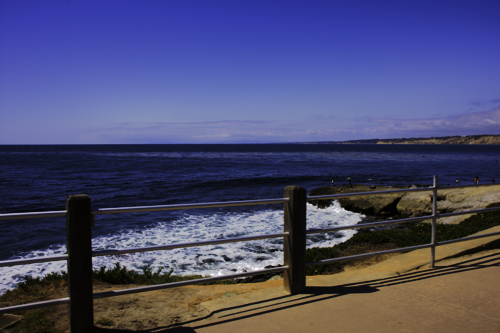}\\
		\scriptsize{White-box}
	\end{minipage}
	\begin{minipage}[t]{0.18\textwidth}
		\centering
		\includegraphics[width=3.25cm]{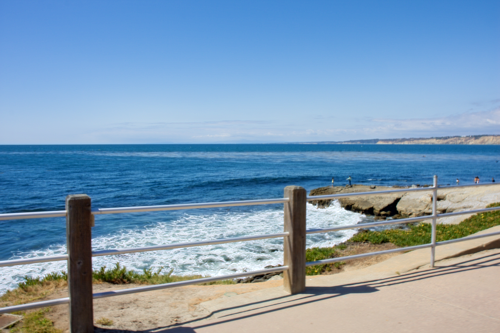}\\
		\scriptsize{DPE}
	\end{minipage}
	\begin{minipage}[t]{0.18\textwidth}
		\centering
		\includegraphics[width=3.25cm]{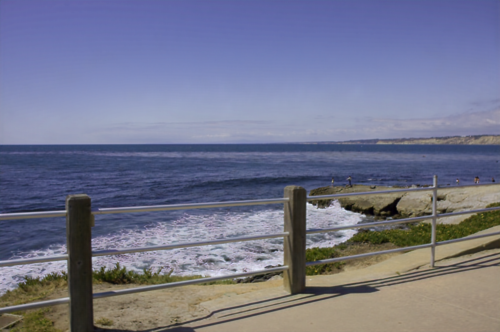}\\
		\scriptsize{MIRNet}
	\end{minipage}\\
	
	\begin{minipage}[t]{0.18\textwidth}
		\centering
		\includegraphics[width=3.25cm]{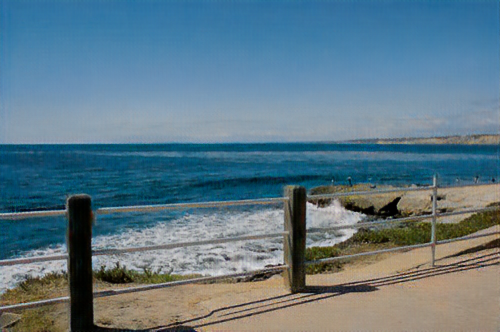}\\
		\scriptsize{Pix2Pix}
	\end{minipage}
	\begin{minipage}[t]{0.18\textwidth}
		\centering
		\includegraphics[width=3.25cm]{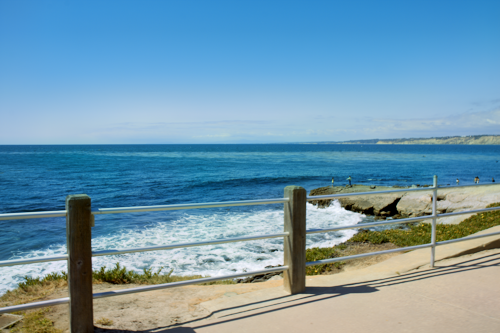}\\
		\scriptsize{HDRNet}
	\end{minipage}
	\begin{minipage}[t]{0.18\textwidth}
		\centering
		\includegraphics[width=3.25cm]{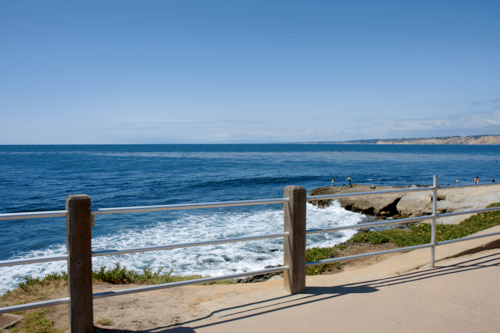}\\
		\scriptsize{3D-LUT}
	\end{minipage}
	\begin{minipage}[t]{0.18\textwidth}
		\centering
		\includegraphics[width=3.25cm]{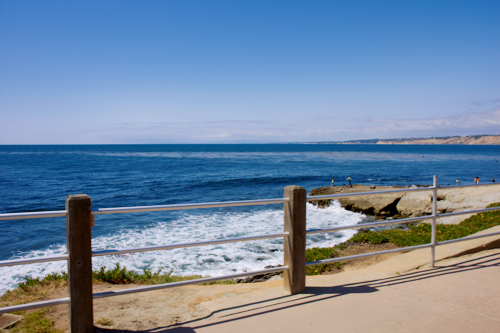}\\
		\scriptsize{CSRNet (ours)}
	\end{minipage}
	\begin{minipage}[t]{0.18\textwidth}
		\centering
		\includegraphics[width=3.25cm]{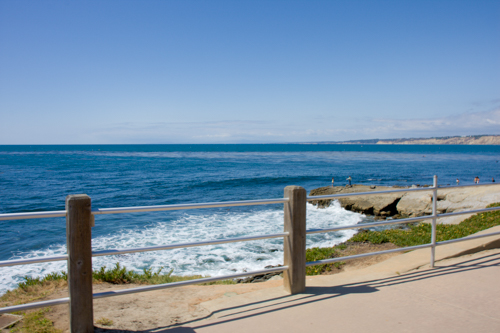}\\
		\scriptsize{GT}
	\end{minipage} \\
	
	\caption{Visual comparison with state-of-the-arts on MIT-Adobe FiveK dataset.}
	\label{fig:main}
\end{figure*}

\subsection{Comparison with State-of-the-art Methods}
To reveal the effectiveness of our method, we compare CSRNet with eight state-of-the-art methods: DUPE \cite{Underexposed}, HDRNet \cite{hdrnet}, DPE \cite{DPE}, MIRNet \cite{MIRNet}, 3D-LUT \cite{3dlut}, White-Box \cite{white-box}, Distort-and-Recover \cite{Distort} and Pix2Pix \cite{pix2pix}\footnote{Pix2Pix uses conditional generative adversarial networks to achieve image-to-image translation and is also applicable to image enhancement problem.}. These methods are all renowned and representative ones in photo retouching, image enhancement or image translation. {For White-Box, DUPE, DPE and MIRNet, we directly use their released pretrained models to test, since their training codes are unavailable. For HDRNet, Distort-and-Recover, 3D-LUT and Pix2Pix, we re-train their models based on their public implementations on the same training dataset as ours. Note that since the training codes of DPE is not yet accessible and their released model is trained on another input version of MIT-Adobe FiveK, we additionally train our model on the same input version for fair comparison.}

\setlength{\tabcolsep}{8pt}
\begin{table*}
	\begin{center}
		\caption{Quantitative comparison with
			state-of-the-art methods on MIT-Adobe FiveK dataset (expert C).}
		\label{table:main}
		\renewcommand{\arraystretch}{1.4}
		\begin{tabular}{lcccccl}
			\hline\noalign{\smallskip}
			Method & Running Time & GFLOPs & PSNR$\uparrow$ & SSIM$\uparrow$ & $\Delta E$ $\downarrow$ & \#params\\
			\noalign{\smallskip}
			\hline
			\noalign{\smallskip}
			Input  & -- -- & -- -- & 10.85 & 0.486 & 34.00 & -- --\\
			White-Box \cite{white-box} & 1028.91ms  & -- -- & 18.59 & 0.797 & 17.42 & 8,561,762\\
			Distort-and-Recover \cite{Distort} & 4063.35ms & -- -- & 19.50 & 0.802 & 15.48 &  259,263,320\\
			HDRNet \cite{hdrnet} & 6.03ms  & 10.60 & 22.65 & 0.880 & 11.83 & 482,080\\
			DUPE \cite{Underexposed} & 8.47ms  & 44.26 & 20.22 & 0.829 & 16.63 & 998,816\\
			MIRNet \cite{MIRNet} & 252.60ms  & 1691.75 & 19.37 & 0.806 & 16.51 & 31,787,419\\
			Pix2Pix \cite{pix2pix} & 181.98ms  & 174.43 & 21.40 & 0.747 & 13.27 & 11,383,427\\
			3D-LUT \cite{3dlut} & 1.60ms  & 0.21 & 23.12 & 0.874 & 11.26 & 593,516\\
			\textbf{CSRNet (ours)} & 1.92ms  & 1.46 & \textbf{23.86} & \textbf{0.897} & \textbf{10.57} & \textbf{36,489}\\
			
			\hline
			\noalign{\smallskip}
			DPE \cite{DPE} & 17.73ms  & 20.75 & 23.76 & 0.881 & 10.60 & 3,335,395\\
			\textbf{CSRNet (ours)} & 1.92ms  & 1.46 & \textbf{24.37} & \textbf{0.902} & \textbf{9.52} & \textbf{36,489}\\
			\hline
		\end{tabular}
	\end{center}
\end{table*}

\textbf{Quantitative Comparison.} We compare CSRNet with state-of-the-art methods in terms of PSNR, SSIM and the Mean L2 error in CIE L*a*b* space {($\Delta E$)}. As we can see from Table~\ref{table:main}, the proposed CSRNet outperforms all the previous state-of-the-art methods by a large margin with the fewest parameters (36,489). Specifically, White-Box and Distort-and-Recover are reinforcement-learning-based methods, which require over millions of parameters but achieve unsatisfactory quantitative results. This is because they are not directly supervised by the ground truth image. HDRNet and DUPE solve the color enhancement problem by estimating the illumination map and require relatively less parameters (less than one million). Since the released model of DUPE is trained for under-exposure images, we can also refer to the result (23.04dB) provided in their paper. Pix2Pix and DPE both utilize the generative adversarial networks and perform well quantitatively. However, they contain much more parameters than CSRNet. Under the same experimental setting, CSRNet outperforms DPE and Pix2Pix in all three metrics with much fewer parameters. {With a cheap $1 \times1$ convolution operation, our method can already achieve better performance with the help of the condition vector. However, with more parameters, the performance could still improve. Experimetns have shown the powerful capability of the condition network for learning global representaions. It is assumed that if other methods are equipped with similar condition network, their performance might be further improved as well.} 3D-LUT is a recent novel method which learns the image-adaptive 3D lookup tables for image enhancement. It achieves good quantitative results with relatively fewer parameters. However, the design of learning lookup tables limits its application. It is only applicable for global photo retouching and it cannot easily be extended to local enhancement tasks. Instead, the proposed CSRNet is a rather flexible framework that can be easily extended to other local enhancement tasks, as detailed in Section \ref{sec:local}.

\textbf{Running Time and FLOPs Comparison.} Ascribing to the specialized design of the base/condition network and the utilization of $1 \times 1$ convolution, CSRNet enjoys a very fast and efficient inference speed at 1.92ms per image (MIT-Adobe FiveK dataset). From Table \ref{table:main}, we can observe that RL-based methods are quite time-consuming. HDRNet \cite{hdrnet} is proposed for real-time image enhancement, while our method is nearly three times faster than HDRNet. 3D-LUT \cite{3dlut} yields the speed of 1.60ms, due to its much lower memory cost. It learns three adaptive 3D lookup tables, which are directly used to map input RGB values to output RGB values. However, it requires more trainable parameters to learn such lookup tables, which are 16 times more than ours. Moreover, 3D-LUT can only be used for global retouching but cannot adapt to local enhancement tasks. {HDRNet, DUPE, DPE, White-Box and Distort-and-recover use TensorFlow framework; Pix2Pix, MIRNet, 3D-LUT and our CSRNet use PyTorch framework. The adopted GPU device is NVIDIA GeForce RTX 2080 Ti. The running time excludes device warm-up time and model preload time.} {In addition, we also compare the FLOPs (floating point operations) of different methods (assume the input image size is $480 \times 480$). As can be seen, CSRNet only have 1.46 GFLOPs, achieving the second best FLOPs with the fewest parameters. In summary, CSRNet is very lightweight and efficient, which is significant for deployment in real applications.}

\textbf{Visual Comparison.}
The results of visual comparison are shown in Figure \ref{fig:main}. The input images from the MIT-Adobe FiveK dataset are generally under low-light condition. Distort-and-recover tends to generate over-exposure output. It seems that White-box and DPE only increase the brightness but fail to modify the original tone, which is oversaturated. The outputs of MIRNet tend to be dark and unsaturated. The enhanced images obtained by Pix2Pix contain artifacts (more visual examples are shown in the supplementary materials). HDRNet outputs images with unnatural color in some regions (e.g. green color on the face and messy color in the flower). The results of 3D-LUT may contain color contaminations in the white sky areas, especially the example of pink flower in Figure \ref{fig:main}. DPE also produces images with color contaminations in the background color. The background color of the pink flower image is little blueish; the overall tone of the house image is blue as well. In conclusion, our method is able to generate more realistic images among all methods.


\begin{figure}[t]
	\centering
	\includegraphics[width=7.5cm]{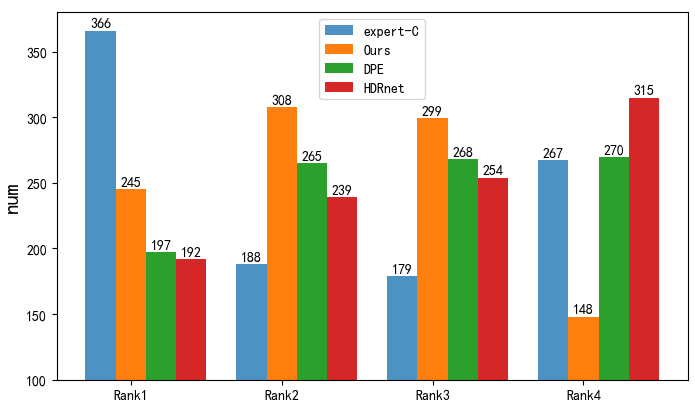}
	\caption{Ranking results of user study. Rank 1 means the best visual quality. Our method is favored by users in most cases.}
	\label{fig:user_study}
\end{figure}

\textbf{User Study.}
We have conducted a user study with 20 participants for subjective evaluation. The participants are asked to rank four retouched image versions (HDRNet \cite{hdrnet}, DPE \cite{DPE}, expert-C (GT) and ours) according to the aesthetic visual quality. 50 images are randomly selected from the testing set and are shown to each participant. 4 retouched versions are displayed on the screen in random order. Users are asked to pay attention to whether the color is vivid, whether there are artifacts and whether the local color is harmonious. Since HDRNet and DPE are representative model-based and GAN-based methods, respectively, we choose them to make the comparison. As suggested in Figure \ref{fig:user_study}, our results achieve better visual ranking against HDRNet and DPE with 553 images ranked first and second. 245 images of our method ranked first, second only to expert C; and 308 images are ranked second, ahead of other methods. Note that, in MIT-Adobe FiveK dataset \cite{mit-adobe}, some of the GT images seem to be darker, due to the retoucher's personal stylistic preference. However, in practice, we find that the users tend to prefer images that contain brighter and vivid color. Hence, in some cases, GT images ranked the last.

{
\subsection{Exploration on Base Network}
The base network of our CSRNet contains 3 convolutional layers with kernel size $1\times1$ and channel number $64$. Here, we explore the base network by changing its kernel size and increasing the number of layers. Besides, we remove the condition network to verify whether the base network could fairly deal with image retouching alone. From Table~\ref{table:base_net}, we can observe that the base network cannot solve the image retouching problem well without the condition network. Specifically, when we expand the filter size to $3\times3$ and increase the number of layers to 7, there is only marginal improvement (0.2 dB) in terms of PSNR. Considering the cases with condition network, if we fix the number of layers, and expand the kernel size to $3\time3$, there is no improvement. Therefore, the process of the base network is just pixel independent, which can be achieved by $1\times1$ filters. {Mathematically, the $1 \times 1$ convolution can be viewed as a special case of $3 \times 3$ convolution. Specifically, if the center value of the convolutional kernel is non-zero and the neighboring values are all zeros, then $3 \times 3$ convolution is equivalent to $1 \times 1$ convolution. However, when equipped with the condition network, the performance of $3 \times 3$ convolution is a little inferior to that of $1 \times 1$ convolution. This is because it is hard to obtain an ideal $3 \times 3$ convolution kernel with only the center value non-zero, due to the issue of approximate optimization. This leads to the fact that the network cannot perfectly simulate the pixel-independent operation and the neighboring pixels will inevitably affect the results. That is why the performance of $3 \times 3$ convolution is not equal to $1 \times 1$ convolution in practice.}
If we fix the kernel size to $1\times1$ and increase the number of layers to 7, the performance improves a little bit (0.08 dB). Since more layers require more parameters, we adopt a lightweight architecture with only three layers.
}

\begin{table}[htbp]
	\begin{center}
		\caption{{Results of ablation study for the base network.}}
		\label{table:base_net}
		\small
		\renewcommand{\arraystretch}{1.4}
		\setlength{\tabcolsep}{2.1pt}
		\begin{tabular}{ccccl}
			\hline\noalign{\smallskip}
			&Base layers& Base kernel size& PSNR  & \#params\\
			\noalign{\smallskip}
			\hline
			\noalign{\smallskip}
			w/o condition &3& $1\times1$ & 20.47 & 4,611 \\
			&3& $3\times3$  & 20.69 & 40,451 \\
			&7& $3\times3$  & 20.67 & 188,163\\
			\hline
			\noalign{\smallskip}
			w condition &3& $1\times1$   & 23.86 &  36,489 \\
			&3& $3\times3$  & 23.75 &  72,329\\
			&5& $1\times1$  & 23.88 &  53,257\\
			&5& $3\times3$  & 23.72 &  154,633\\
			&7& $1\times1$  & 23.94 &  70,025\\
			&7& $3\times3$  & 23.73 &  236,937\\
			\hline
		\end{tabular}
	\end{center}
\end{table}

\subsection{Exploration on Normalization Strategies}\label{sec:explore_normalization}
In Section \ref{sec:normalize}, we introduce the unit normalization (UN) operation to normalize the condition vectors, which can enhance the training stability and improve the performance. We further make explorations on various feature normalization methods: 1) Without any normalization (None). 2) Softmax normalization. 3) Sigmoid normalization. 4) Softmax scaling. 5) Sigmoid scaling. 6) Z-score normalization (standardization). 7) Min-max normalization (Rescaling). 8) The proposed unit normalization (UN). Their formulas are depicted in Table \ref{tab:normalization}.

\begin{table}[htbp]
	\begin{center}
		\centering
		\caption{Exploration on various feature normalization methods. The proposed unit normalization (UN) significant improves the performance. $x_i':$ the transformed feature. $\mu:$ mean value. $\sigma:$ standard deviation.}
		\label{tab:normalization}
		\small
		\renewcommand{\arraystretch}{1.4}
		\setlength{\tabcolsep}{9pt}
		\begin{tabular}{c|c|c}
			\hline
			Normalization/scaling & Formula & PSNR \\ \hline
			None & $x_i'=x_i$ & 23.58 \\ \hline
			Softmax & $x_i'=\frac{e^{x_i}}{\sum_{j=1}e^j}$ & 23.47 \\ \hline
			Sigmoid & $x_i'=\frac{1}{1+e^{-x_i}}$ & 23.62 \\ \hline
			Softmax scaling & $x_i'=\frac{e^{x_i}}{\sum_{j=1}e^j}x_i$ & 23.48 \\ \hline
			Sigmoid scaling & $x_i'=\frac{1}{1+e^{-x_i}}x_i$ & 23.61 \\ \hline
			Z-score & $x_i'=\frac{x_i - \mu}{\sigma}$ & 23.69 \\ \hline
			Min-max & $x_i'=\frac{x_i - \min (x)}{\max (x) - \min (x)}$ & 23.68 \\ \hline
			UN (proposed) & $x_i'= \frac{\sqrt{N} x_i}{\left \| x \right \|_2}$ & \textbf{23.86} \\ \hline
		\end{tabular}
	\end{center}
	
\end{table}

\begin{figure}[htbp]
	\centering
	\begin{minipage}[t]{0.325\linewidth}
		\centering
		\includegraphics[width=0.98\linewidth,height=0.7\linewidth]{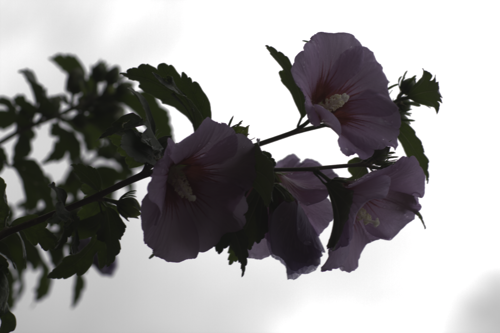}\\
		\scriptsize{input}
	\end{minipage}
	\begin{minipage}[t]{0.325\linewidth}
		\centering
		\includegraphics[width=0.98\linewidth,height=0.7\linewidth]{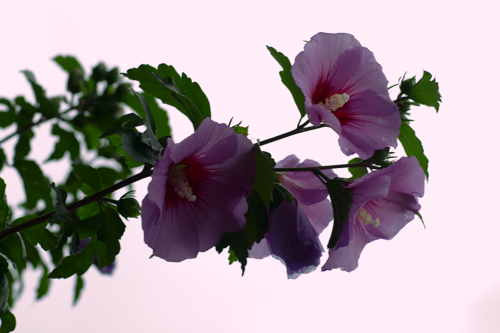}\\
		\scriptsize{None}
	\end{minipage}
	\begin{minipage}[t]{0.325\linewidth}
		\centering
		\includegraphics[width=0.98\linewidth,height=0.7\linewidth]{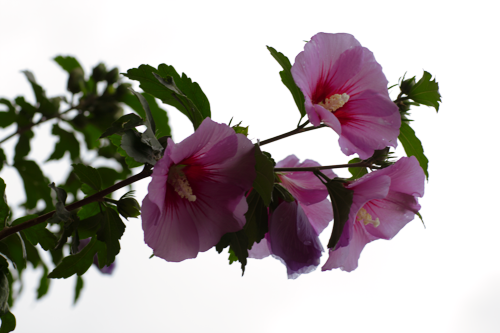}\\
		\scriptsize{Softmax}
	\end{minipage}\\
	
	\begin{minipage}[t]{0.325\linewidth}
		\centering
		\includegraphics[width=0.98\linewidth,height=0.7\linewidth]{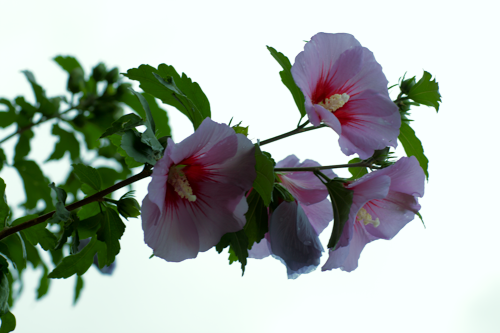}\\
		\scriptsize{Sigmoid}
	\end{minipage}
	\begin{minipage}[t]{0.325\linewidth}
		\centering
		\includegraphics[width=0.98\linewidth,height=0.7\linewidth]{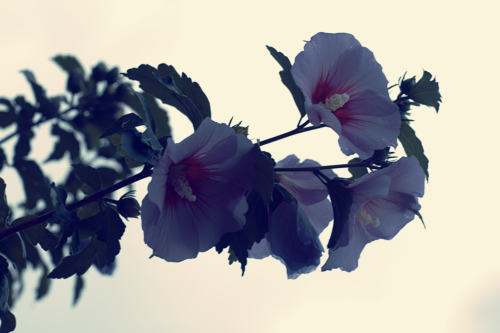}\\
		\scriptsize{Softmax scaling}
	\end{minipage}
	\begin{minipage}[t]{0.325\linewidth}
		\centering
		\includegraphics[width=0.98\linewidth,height=0.7\linewidth]{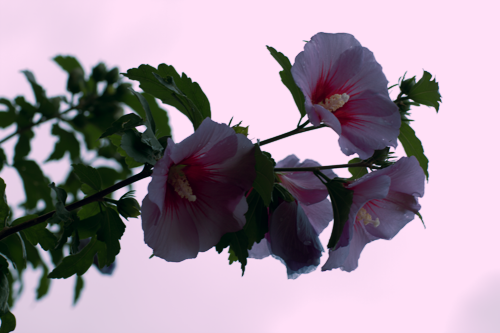}\\
		\scriptsize{Sigmoid scaling}
	\end{minipage}\\
	
	\begin{minipage}[t]{0.325\linewidth}
		\centering
		\includegraphics[width=0.98\linewidth,height=0.7\linewidth]{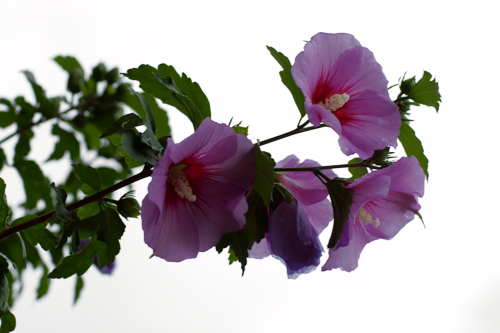}\\
		\scriptsize{Z-score}
	\end{minipage}
	\begin{minipage}[t]{0.325\linewidth}
		\centering
		\includegraphics[width=0.98\linewidth,height=0.7\linewidth]{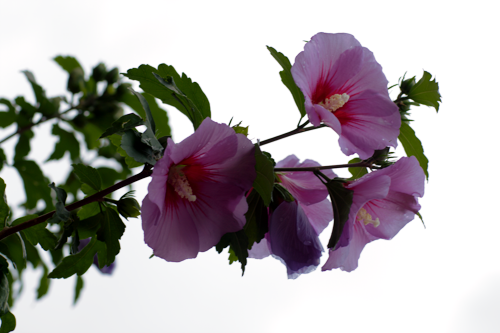}\\
		\scriptsize{Min-max}
	\end{minipage}
	\begin{minipage}[t]{0.325\linewidth}
		\centering
		\includegraphics[width=0.98\linewidth,height=0.7\linewidth]{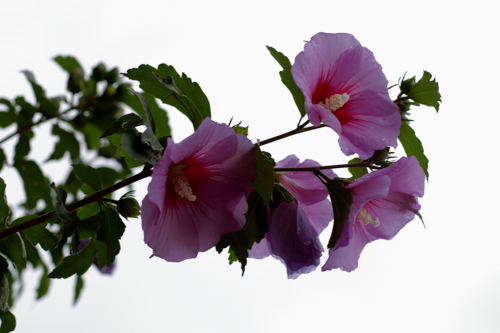}\\
		\scriptsize{UN}
	\end{minipage}\\
	
	\vspace{5pt}
	
	\begin{minipage}[t]{0.325\linewidth}
		\centering
		\includegraphics[width=0.98\linewidth,height=0.8\linewidth]{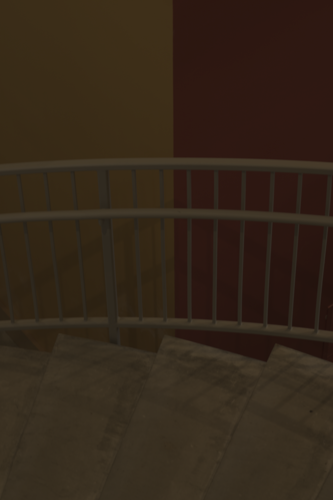}\\
		\scriptsize{input}
	\end{minipage}
	\begin{minipage}[t]{0.325\linewidth}
		\centering
		\includegraphics[width=0.98\linewidth,height=0.8\linewidth]{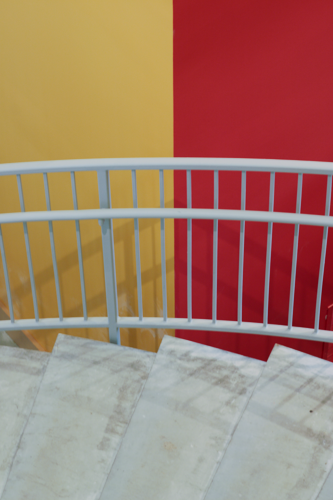}\\
		\scriptsize{None}
	\end{minipage}
	\begin{minipage}[t]{0.325\linewidth}
		\centering
		\includegraphics[width=0.98\linewidth,height=0.8\linewidth]{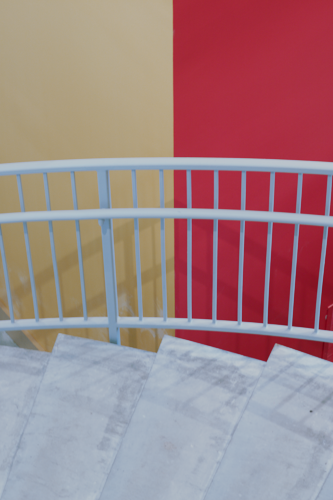}\\
		\scriptsize{Softmax}
	\end{minipage}\\
	
	\begin{minipage}[t]{0.325\linewidth}
		\centering
		\includegraphics[width=0.98\linewidth,height=0.8\linewidth]{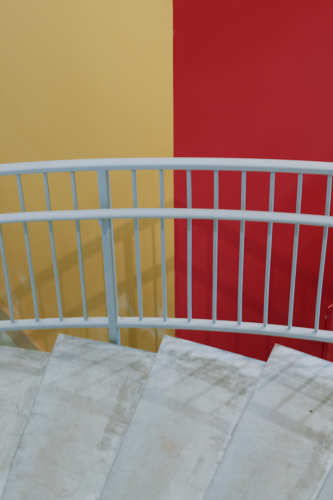}\\
		\scriptsize{Sigmoid}
	\end{minipage}
	\begin{minipage}[t]{0.325\linewidth}
		\centering
		\includegraphics[width=0.98\linewidth,height=0.8\linewidth]{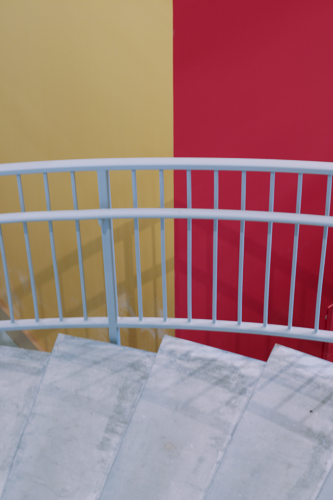}\\
		\scriptsize{Softmax scaling}
	\end{minipage}
	\begin{minipage}[t]{0.325\linewidth}
		\centering
		\includegraphics[width=0.98\linewidth,height=0.8\linewidth]{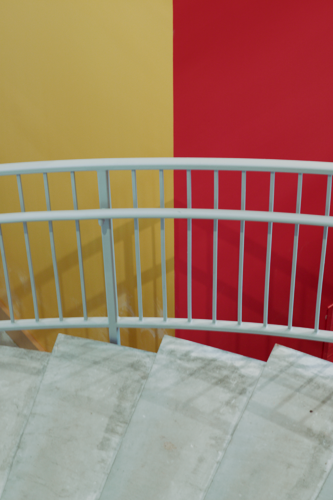}\\
		\scriptsize{Sigmoid scaling}
	\end{minipage}\\
	
	\begin{minipage}[t]{0.325\linewidth}
		\centering
		\includegraphics[width=0.98\linewidth,height=0.8\linewidth]{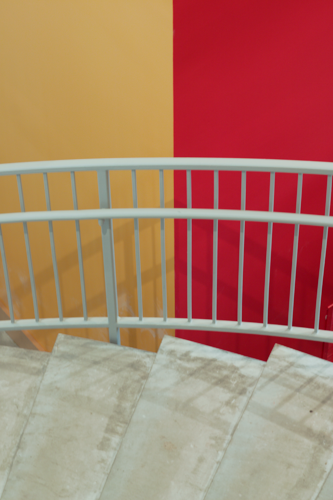}\\
		\scriptsize{Z-score}
	\end{minipage}
	\begin{minipage}[t]{0.325\linewidth}
		\centering
		\includegraphics[width=0.98\linewidth,height=0.8\linewidth]{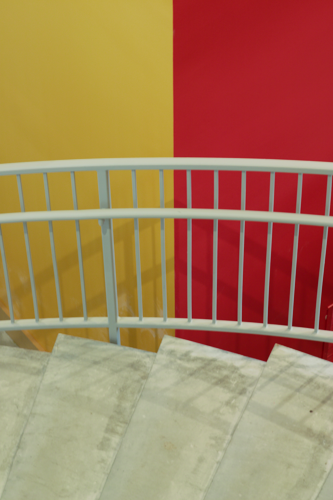}\\
		\scriptsize{Min-max}
	\end{minipage}
	\begin{minipage}[t]{0.325\linewidth}
		\centering
		\includegraphics[width=0.98\linewidth,height=0.8\linewidth]{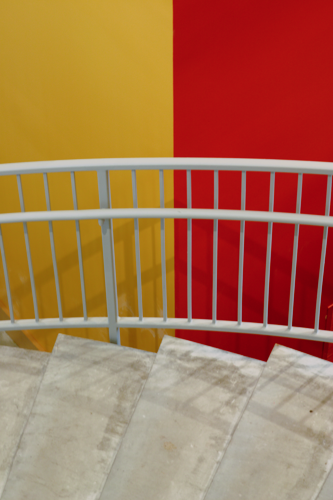}\\
		\scriptsize{UN}
	\end{minipage}\\
	
	\caption{Visual comparison among various normalization or scaling operations on the condition vector. In the first example (1st row -- 3rd row), the background color of the input image is nearly all white (an extreme value). Without UN, CSRNet tends to confuse the foreground object color with the background color, and outputs a pink background. By introducing UN, the problem can be well solved. In the second example (4th row -- 6th row), adopting UN can produce more vivid and saturated colors (yellow wall).}
	\label{fig:normalization}
\end{figure}

\begin{figure*}[htbp]
	\centering
	\begin{minipage}[htbp]{0.48\linewidth}
		\centering
		\includegraphics[width=1\linewidth,height=0.3\linewidth]{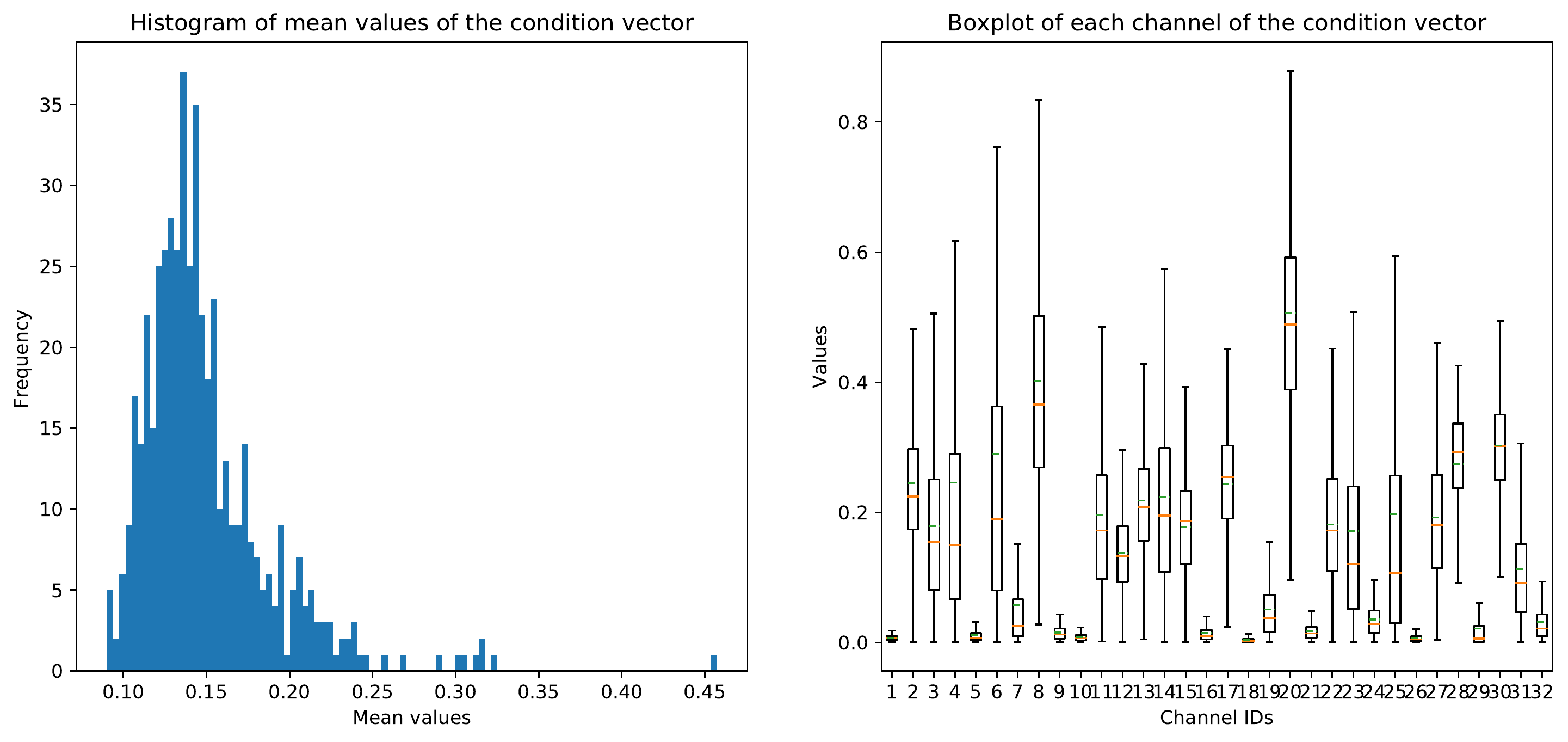}\\
		\scriptsize None
	\end{minipage}
	\begin{minipage}[htbp]{0.48\linewidth}
		\centering
		\includegraphics[width=1\linewidth,height=0.3\linewidth]{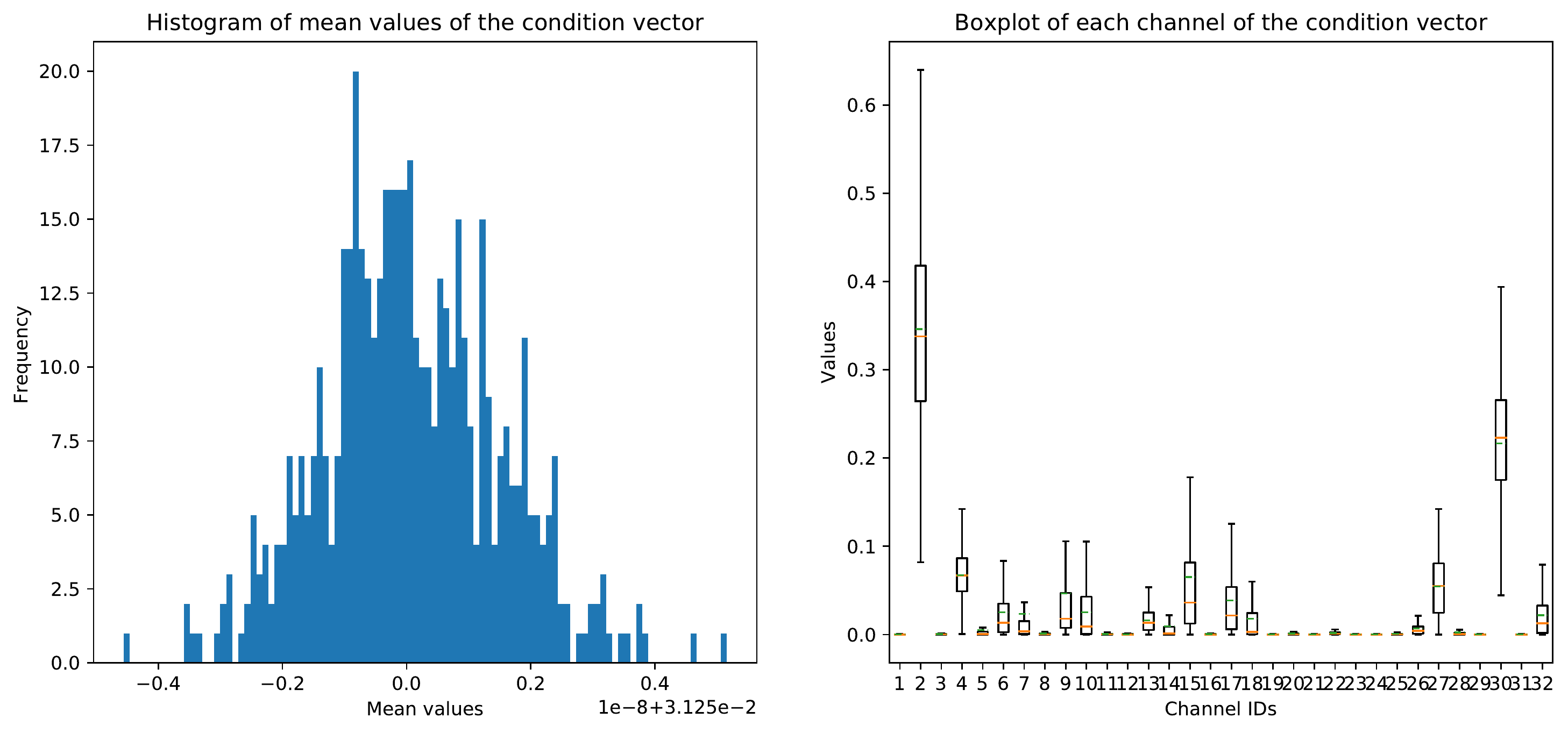}\\
		\scriptsize Softmax
	\end{minipage}\\
	\begin{minipage}[htbp]{0.48\linewidth}
		\centering
		\includegraphics[width=1\linewidth,height=0.3\linewidth]{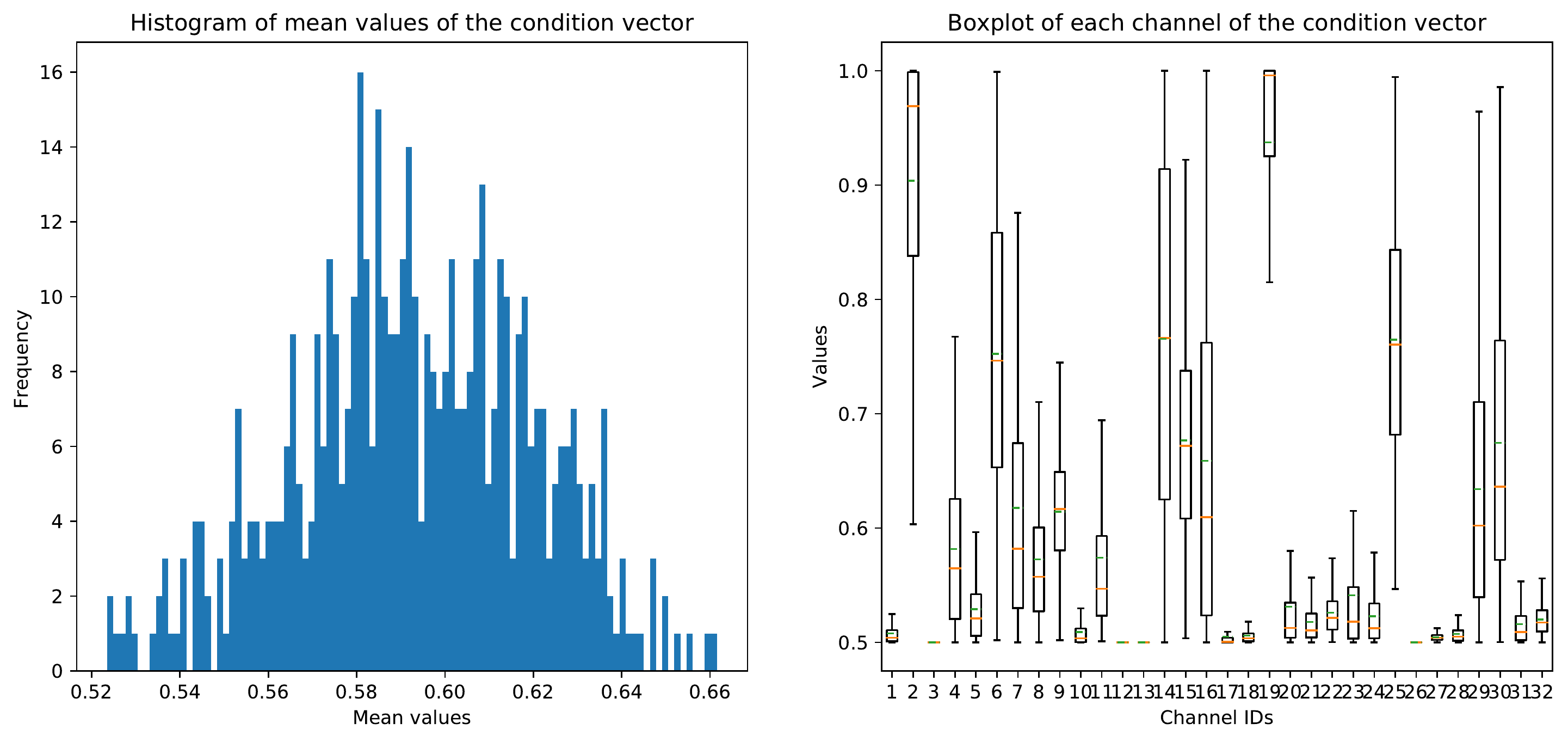}\\
		\scriptsize Sigmoid
	\end{minipage}
	\begin{minipage}[htbp]{0.48\linewidth}
		\centering
		\includegraphics[width=1\linewidth,height=0.3\linewidth]{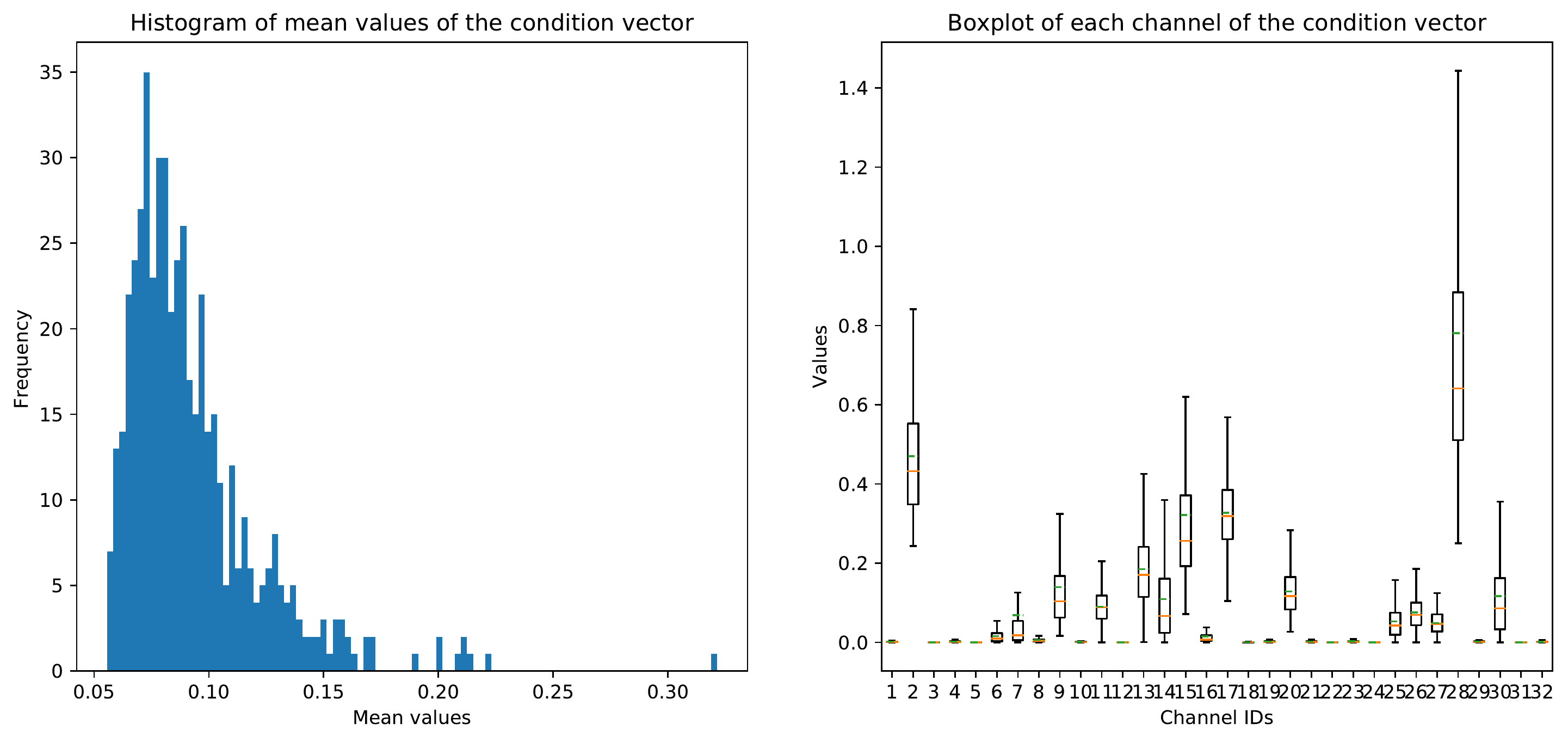}\\
		\scriptsize Softmax scaling
	\end{minipage}\\
	
	\begin{minipage}[htbp]{0.48\linewidth}
		\centering
		\includegraphics[width=1\linewidth,height=0.3\linewidth]{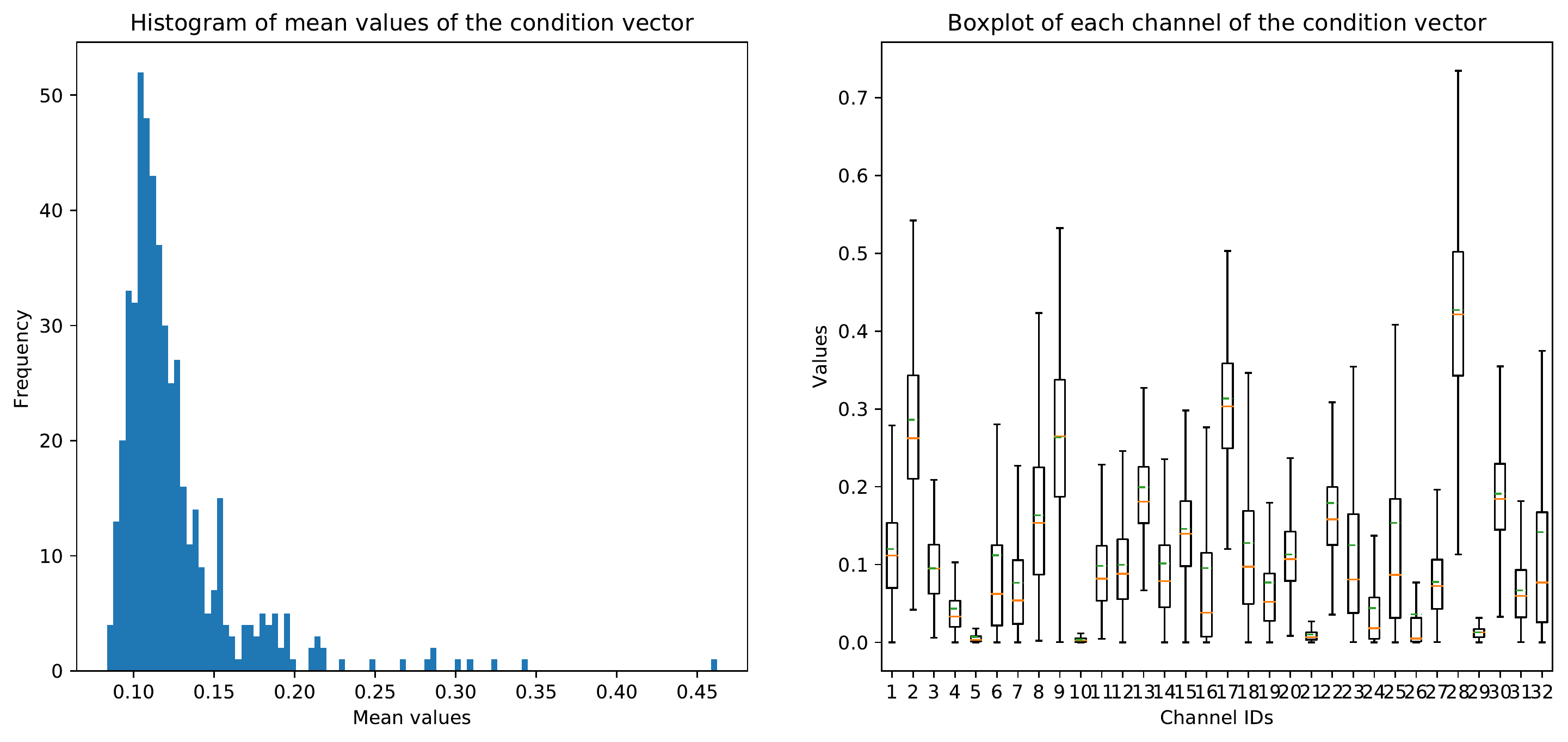}\\
		\scriptsize Sigmoid scaling
	\end{minipage}
	\begin{minipage}[htbp]{0.48\linewidth}
		\centering
		\includegraphics[width=1\linewidth,height=0.3\linewidth]{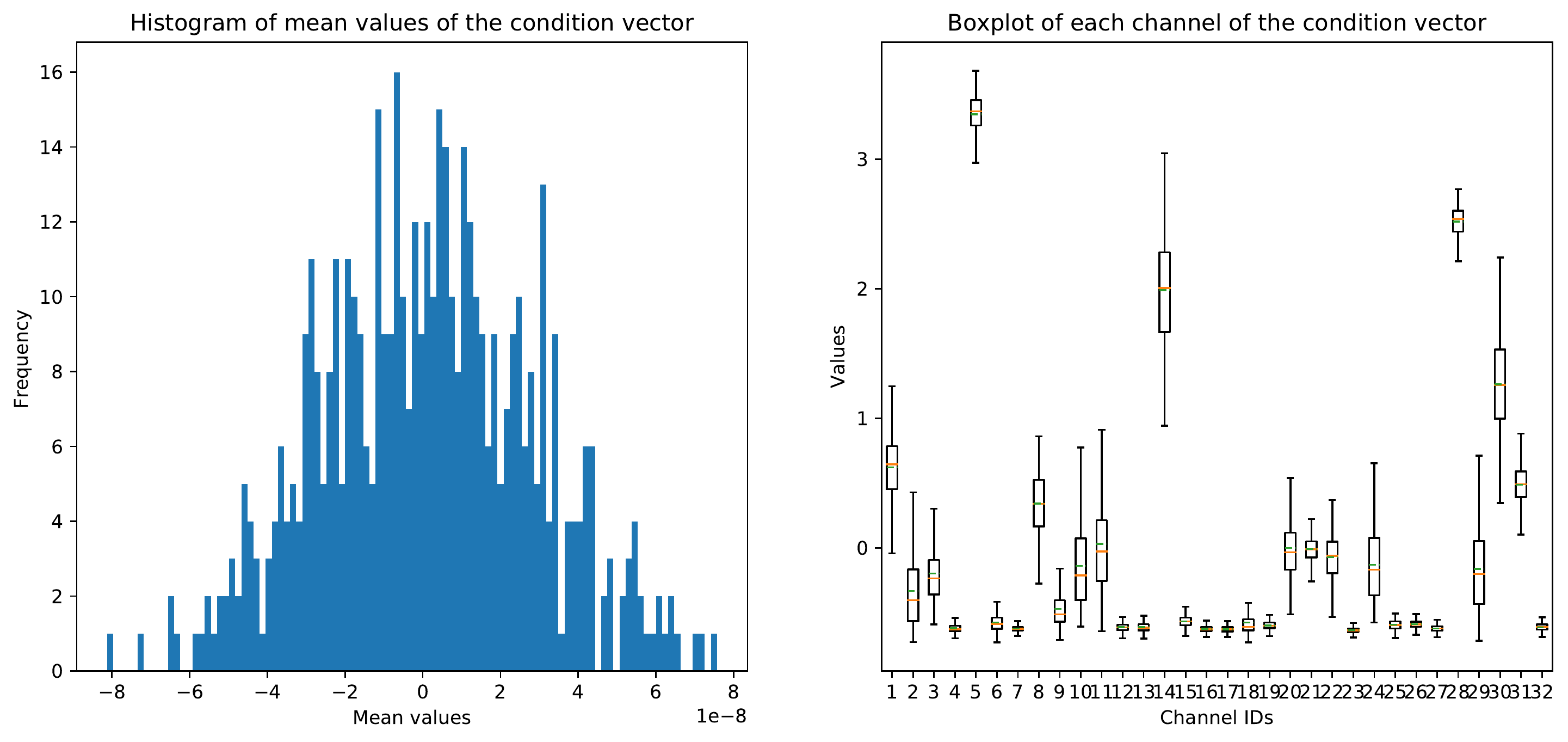}\\
		\scriptsize Z-score
	\end{minipage}
	\begin{minipage}[htbp]{0.48\linewidth}
		\centering
		\includegraphics[width=1\linewidth,height=0.3\linewidth]{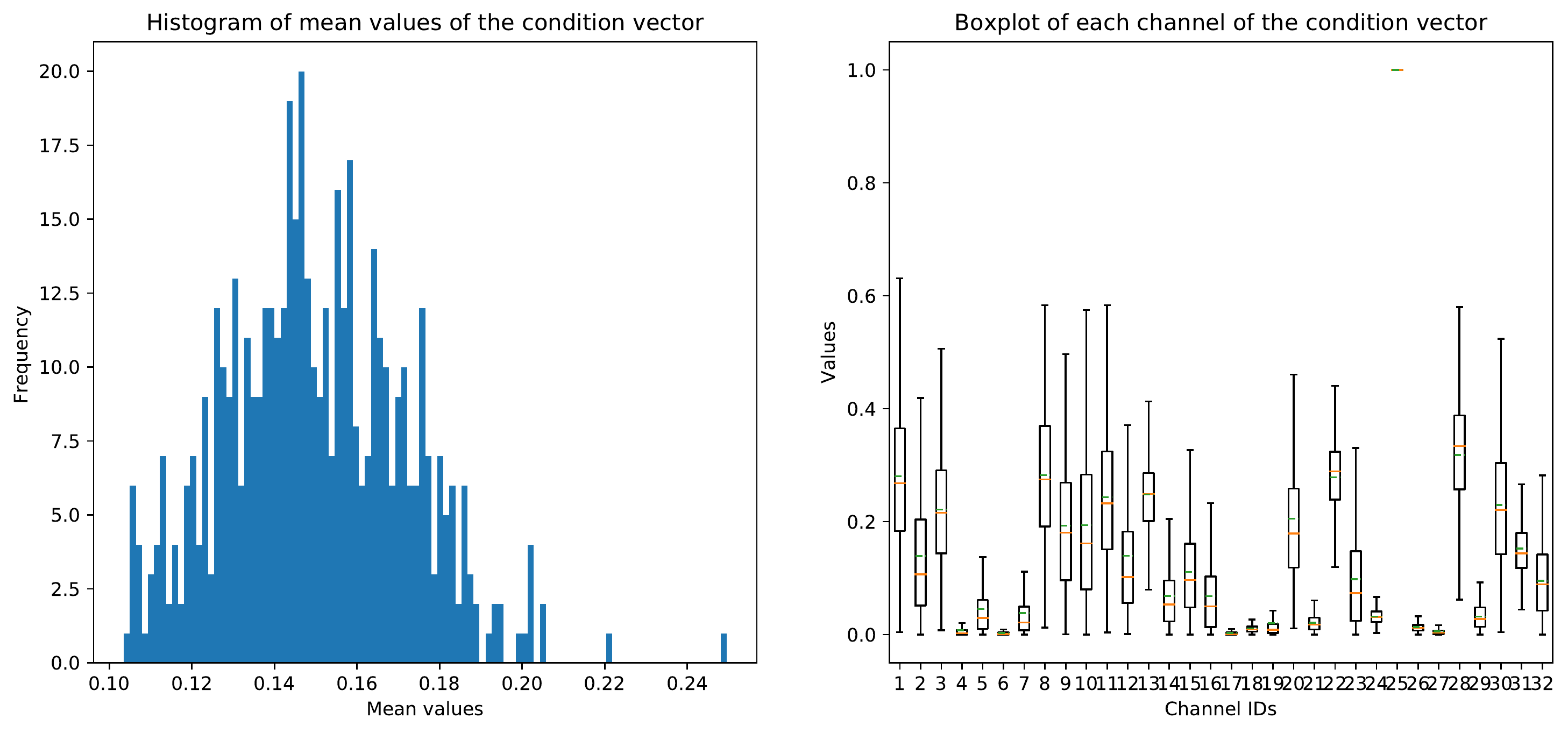}\\
		\scriptsize Min-max
	\end{minipage}
	\begin{minipage}[htbp]{0.48\linewidth}
		\centering
		\includegraphics[width=1\linewidth,height=0.4\linewidth]{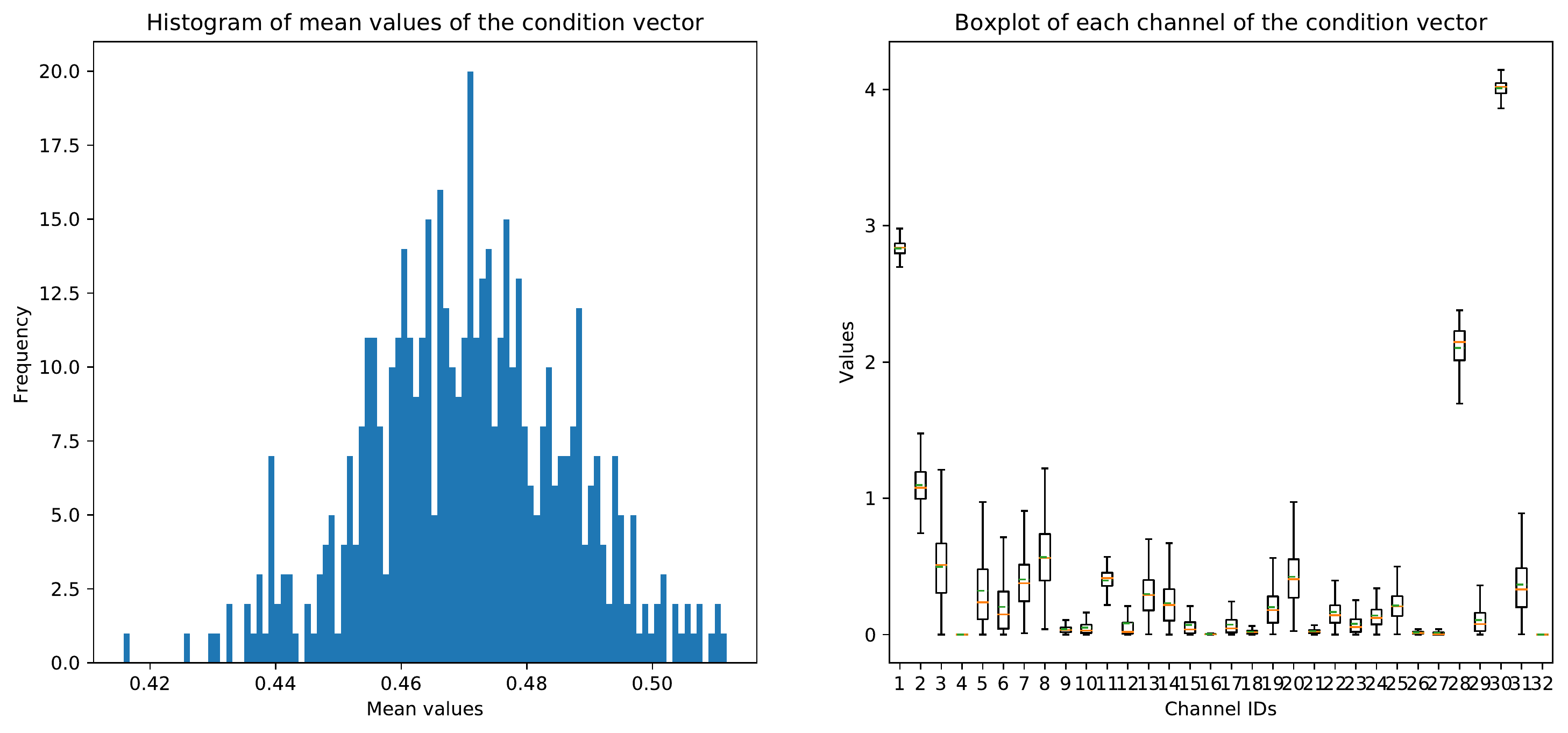}\\
		\scriptsize UN (proposed)
	\end{minipage}\\

	\caption{{Distributions of the condition vector with different normalization/scaling methods. \textbf{Left part:} Histogram of mean values of the condition vector. \textbf{Right part:} Boxplot of each channel of the condition vector.}}
	\label{fig:distribution}
\end{figure*}

As shown in Table \ref{tab:normalization}, adopting proper normalization operations can help improve the performance. Sigmoid normalization, Sigmoid scaling, Z-score, Min-max and UN all bring PSNR improvement. Specifically, the quantitative results of Z-score and Min-max normalization are similar, which improve the PSNR value by 0.11 and 0.10 dB. Nevertheless, Softmax normalization and Softmax scaling could cause obvious performance drop. {Softmax operation will enhance the large values among all the elements and suppress the small values.} Z-score and Min-max normalization perform similarly, since their formulations are similar with a shift and a scale operation. The proposed UN does not include a shift operation, but a scaling operation with $\sqrt{N}$-sphere unitization. The proposed UN can restrict the value ranges, largely alleviating outliers in each channel, making the values more concentrated. Among all these operations, the proposed UN operation achieves the highest PSNR value, improving the PSNR by 0.28 dB.

Qualitatively, the visual results of various feature normalization/scaling operations are also different. As shown in Figure \ref{fig:normalization}, for some special cases, if the condition vector is not well normalized, the model will confuse the foreground object color with the background color and output a wrong background color. Such a problem can be also observed in other methods, like DPE, MIRNet, HDRNet and 3D-LUT. By adopting appropriate normalization operations in our method, this issue can be well solved. Further, comparing with other common normalization strategy, the proposed UN can generate more vivid and saturated retouched results. In the second example in Figure \ref{fig:normalization}, the output produced by Softmax is somewhat whitish with unsaturated and low-contrast color, while UN can generate a more saturated and vibrant yellow wall. In summary, both quantitative and qualitative results have demonstrated the effectiveness of the proposed UN operation.

{
Further, we compute the average mean of the prediceted condition vectors over 500 testing images with different normalization/scaling methods and plot the histograms. This can show the distribution of the mean values over 500 samples. Besides, we draw the boxplots of each channel of the condition vector (32 in total) to display the distribution of each channel. The visualization results are shown in Figure \ref{fig:distribution}. If we do not apply any normalization, the distribution of the mean values of the condition vector is non-Gaussian-like with skewed bias. After Softmax, most channel values are suppressed to zero, and the other values still maintain a large jitter range. The results reveal that Softmax could degrade the performance, since it makes most channel values near zero. After applying UN normalization, the distribution will be Gaussian-like. As for each channel of the condition vector, the unnormalized original values vary a lot in a wide range and contain several outliers: the variance within each channel is large, leading to unstable outliers when extreme cases occur. The proposed UN can restrict the value ranges, largely alleviating outliers in each channel, making the values more concentrated. As shown in Figure \ref{fig:distribution}, the condition vectors become more compact.
}

\section{Learning Stylistic Local Effects}
To demonstrate the effectiveness of CSRNet-L for local adjustments, we conduct experiments with four stylistic local effects. It is shown that our extended method can attain comparable performance with other specially designed networks.

\textbf{Datasets.} We evaluate the performance of our method on four local enhancement tasks:
\begin{itemize}
	\item Fast Local Laplacian Filter \cite{fastllf}. A multi-scale operator for edge-preserving detail enhancement. We apply this operator on images retouched by expert C of the MIT-Adobe FiveK dataset \cite{mit-adobe}.
\end{itemize}

The following three datasets are proposed by \cite{automatic}. 115 images from Flickr are selected and retouched by a professional photographer using Photoshop. 70 images were chosen for training and the remaining 45 images for testing. The photographer performed a wide range of operations to adjust the images and created three different stylistic local effects. Here we make a brief introduction to them. 
\begin{itemize}
	\item Foreground Pop-Out. The photographer increased both the contrast and color saturation of foreground salient objects, while suppressed that of the background at the same time. Consequently, the foreground is highlighted and seems to ``pop-out''.
	\item Local Xpro. This effect was produced by generalizing the prevailing ``cross processing'' \footnote{Cross processing (abbreviated to Xpro) is the deliberate processing of photographic film in a chemical solution intended for a different type of film. Cross processed photographs are often characterized by unnatural colors and high contrast.} effect in a local manner. The photographer first isolated different image regions and then applied a series of operations on each region according to the semantic content within that region.
	\item Watercolor. This effect makes the images look like ``watercolor'' painting style which tends to be brighter and less saturated. Please refer to \cite{automatic} for detailed procedures of creating this effect.
\end{itemize}

\begin{table*}[!htbp]
	\begin{center}
		\caption{Quantitative results on local effect datasets. By extending the CSRNet, our CSRNet-L successfully achieves local effect adjustments and obtains competitive performance with much fewer parameters. \textbf{Bold} and \textit{\underline{italic}} indicate the best and the second best performance, respectively.}
		\label{table:local}
		\small
		\renewcommand{\arraystretch}{1.4}
		\setlength{\tabcolsep}{4pt}
		\begin{tabular}{c|cccccccc}
			\hline
			&\multicolumn{2}{c}{LLF}&\multicolumn{2}{c}{Foreground}&\multicolumn{2}{c}{Xpro}&\multicolumn{2}{c}{Watercolor}\\
			&\underline{PSNR}&\underline{SSIM}&\underline{PSNR}&\underline{SSIM}&\underline{PSNR}&\underline{SSIM}&\underline{PSNR}&\underline{SSIM} \\
			
			\hline 
			Input&18.92&0.648&23.31&0.933&18.56&0.925&19.62&0.657\\
			Pix2Pix \cite{pix2pix}&24.76&{0.879}&24.12&0.884&26.85&0.926&22.43&0.739 \\ 
			HDRNet \cite{hdrnet}&{24.94}&0.871&{26.73}&\textit{\underline{0.943}}&\textbf{30.30}&\textbf{0.975}&{23.50}&{0.800} \\
			CAN \cite{can}&\textbf{26.90}&\textit{\underline{0.900}}&\textit{\underline{27.01}}&{0.938}&\textit{\underline{29.72}}&{0.965}&\textit{\underline{24.27}}&\textit{\underline{0.828}} \\
			
			CSRNet-L (3 layers)&\textit{\underline{25.03}}&\textbf{0.904}&\textbf{27.13}&\textbf{0.943}&{29.48}&\textit{\underline{0.968}}&\textbf{24.35}&\textbf{0.832} \\

			\hline
		\end{tabular}
	\end{center}
\end{table*}

\begin{figure}[h]
	\centering
	\begin{minipage}[t]{0.2\textwidth}
		\centering
		\includegraphics[width=3.6cm]{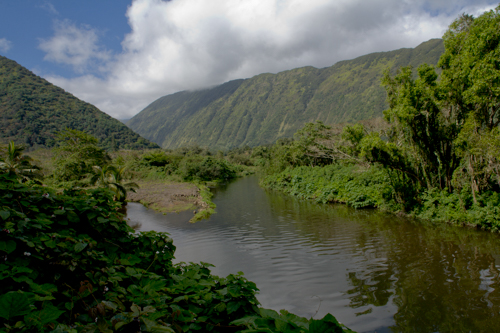}\\
		\scriptsize{input}
	\end{minipage}
	\begin{minipage}[t]{0.2\textwidth}
		\centering
		\includegraphics[width=3.6cm]{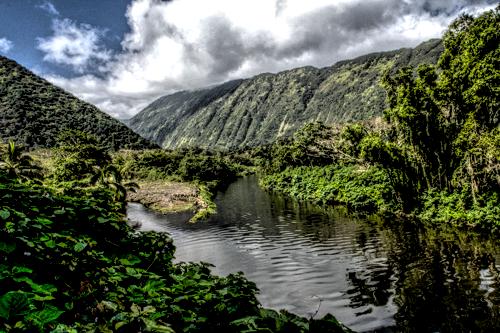}\\
		\scriptsize{GT}
	\end{minipage}
	
	\begin{minipage}[t]{0.2\textwidth}
		\centering
		\includegraphics[width=3.6cm]{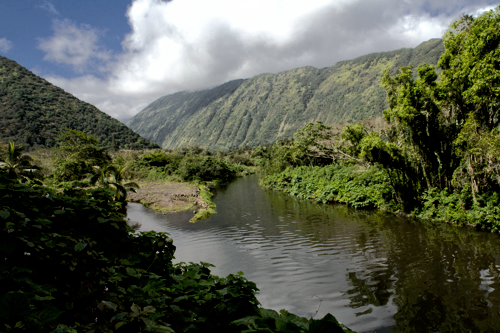}\\
		\scriptsize{CSRNet ($1 \times 1$ + GFM)}
	\end{minipage}
	\begin{minipage}[t]{0.2\textwidth}
		\centering
		\includegraphics[width=3.6cm]{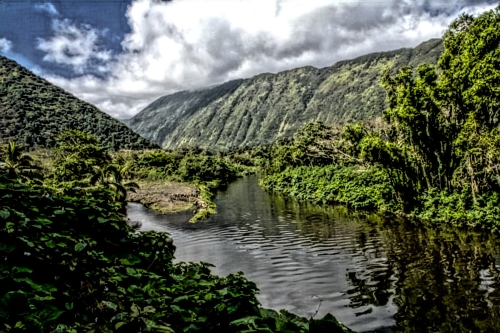}\\
		\scriptsize{CSRNet-L ($3 \times 3$ + SFM)}
	\end{minipage}
	
	\begin{minipage}[t]{0.2\textwidth}
		\centering
		\includegraphics[width=3.6cm]{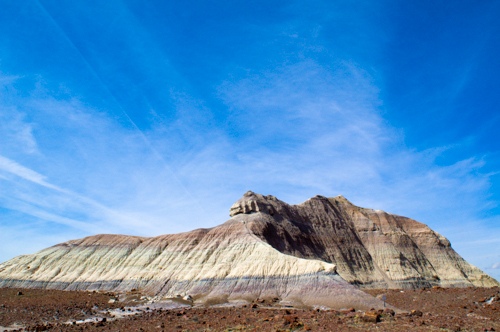}\\
		\scriptsize{input}
	\end{minipage}
	\begin{minipage}[t]{0.2\textwidth}
		\centering
		\includegraphics[width=3.6cm]{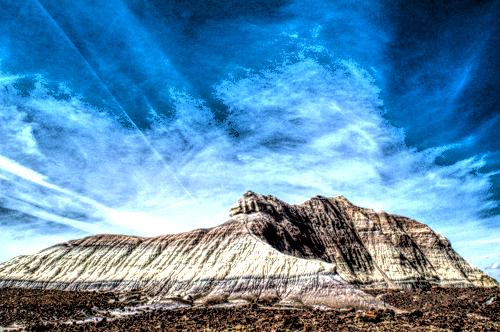}\\
		\scriptsize{GT}
	\end{minipage}
	
	\begin{minipage}[t]{0.2\textwidth}
		\centering
		\includegraphics[width=3.6cm]{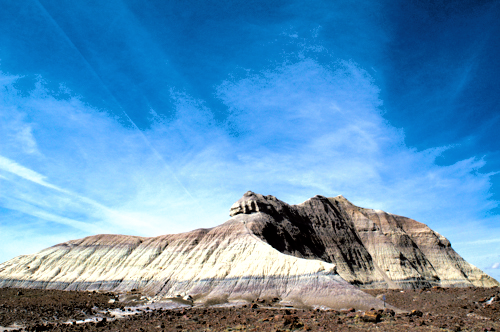}\\
		\scriptsize{CSRNet ($1 \times 1$ + GFM)}
	\end{minipage}
	\begin{minipage}[t]{0.2\textwidth}
		\centering
		\includegraphics[width=3.6cm]{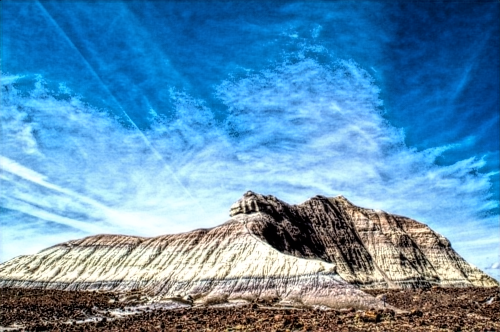}\\
		\scriptsize{CSRNet-L ($3 \times 3$ + SFM)}
	\end{minipage}
	
	\caption{From global to local. By expanding the filter size of the base network and adopting SFM, CSRNet-L can achieve local effects, demonstrating the expansibility and superiority of the proposed framework.}
	\label{fig:csrnet_local}
\end{figure}

\textbf{Implementation details.} The network structure is similar with that in Section \ref{sec:network}, except that we expand the filter size of the base network to $3 \times 3$, and we adopt SFM instead of GFM, as depicted in Figure \ref{fig:frame_csrnetl}. The downsampling operations are removed from the condition network so that the output size is the same as the feature map size in the base network. To achieve local effects, we first train the base network, and then add the condition network for joint training. For training the base network, the learning rate is initialized to $1 \times 10^{-4}$, while the learning is initialized to $1 \times 10^{-5}$ for joint training.  We find this two-stage training strategy can obtain more stable and better performance on local enhancement tasks.

\begin{figure*}[htbp]
	\centering
	\begin{minipage}[t]{0.19\textwidth}
		\centering
		\includegraphics[width=3.4cm,height=2.1cm]{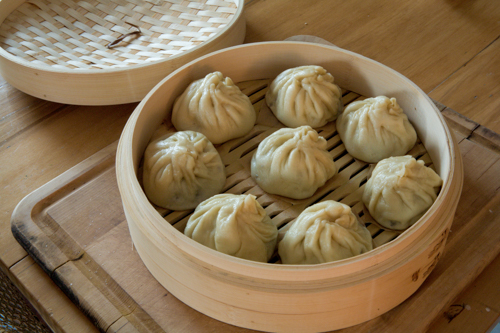}\\
		\scriptsize{(a) input/LLF}
	\end{minipage}
	\begin{minipage}[t]{0.19\textwidth}
		\centering
		\includegraphics[width=3.4cm,height=2.1cm]{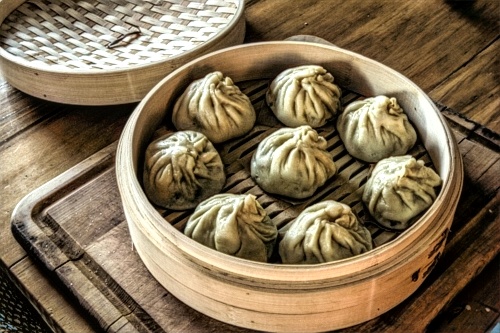}\\
		\scriptsize{CAN}
	\end{minipage}
	\begin{minipage}[t]{0.19\textwidth}
		\centering
		\includegraphics[width=3.4cm,height=2.1cm]{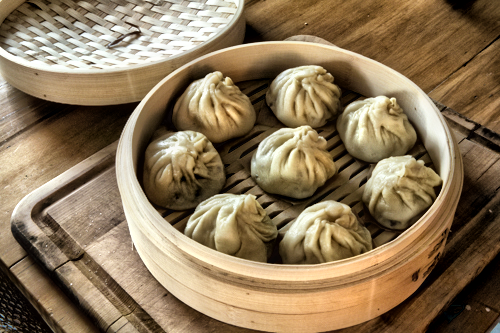}\\
		\scriptsize{HDRNet}
	\end{minipage}
	\begin{minipage}[t]{0.19\textwidth}
		\centering
		\includegraphics[width=3.4cm,height=2.1cm]{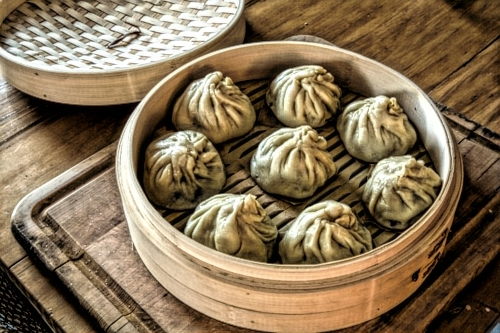}\\
		\scriptsize{CSRNet-L}
	\end{minipage}
	\begin{minipage}[t]{0.19\textwidth}
		\centering
		\includegraphics[width=3.4cm,height=2.1cm]{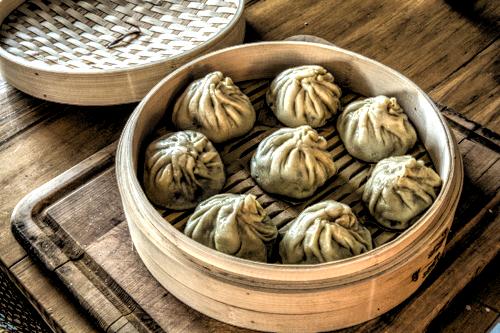}\\
		\scriptsize{GT}
	\end{minipage}
			
	\begin{minipage}[t]{0.19\textwidth}
		\centering
		\includegraphics[width=3.4cm,height=2.1cm]{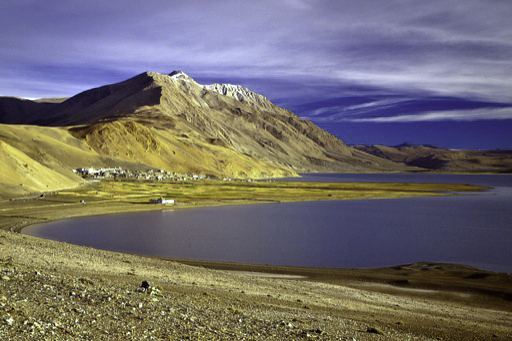}\\
		\scriptsize{(b) input/Foreground Pop-Out}
	\end{minipage}
	\begin{minipage}[t]{0.19\textwidth}
		\centering
		\includegraphics[width=3.4cm,height=2.1cm]{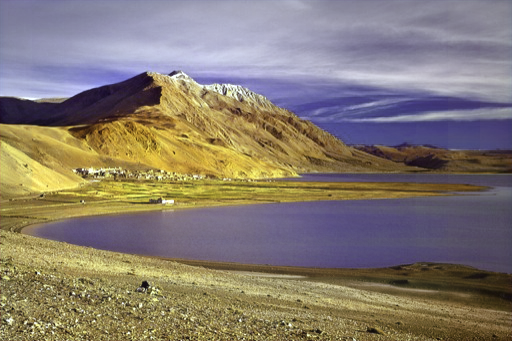}\\
		\scriptsize{CAN}
	\end{minipage}
	\begin{minipage}[t]{0.19\textwidth}
		\centering
		\includegraphics[width=3.4cm,height=2.1cm]{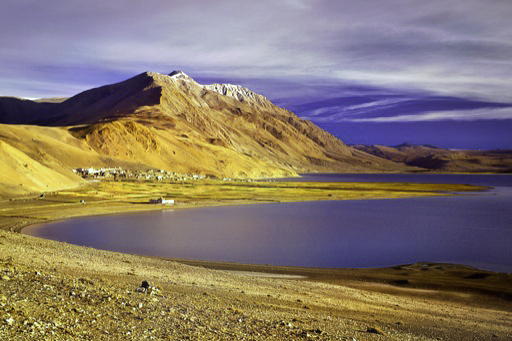}\\
		\scriptsize{HDRNet}
	\end{minipage}
	\begin{minipage}[t]{0.19\textwidth}
		\centering
		\includegraphics[width=3.4cm,height=2.1cm]{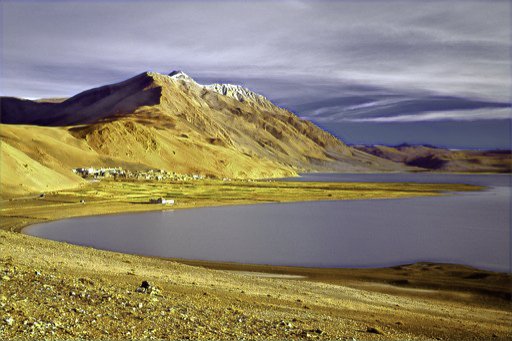}\\
		\scriptsize{CSRNet-L}
	\end{minipage}
	\begin{minipage}[t]{0.19\textwidth}
		\centering
		\includegraphics[width=3.4cm,height=2.1cm]{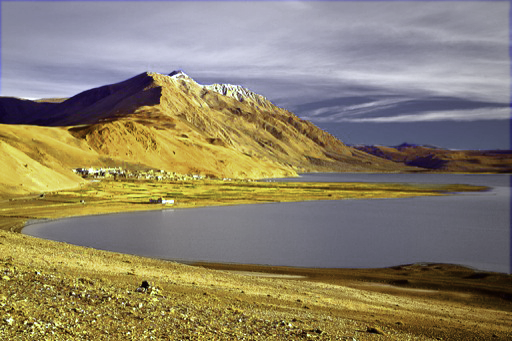}\\
		\scriptsize{GT}
	\end{minipage}
	
	\begin{minipage}[t]{0.19\textwidth}
		\centering
		\includegraphics[width=3.4cm,height=2.1cm]{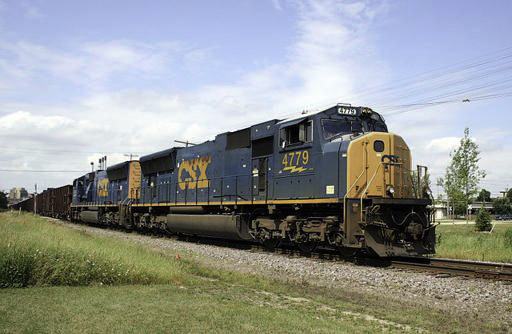}\\
		\scriptsize{(c) input/Local Xpro}
	\end{minipage}
	\begin{minipage}[t]{0.19\textwidth}
		\centering
		\includegraphics[width=3.4cm,height=2.1cm]{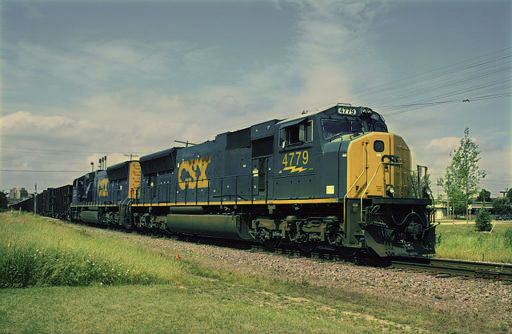}\\
		\scriptsize{CAN}
	\end{minipage}
	\begin{minipage}[t]{0.19\textwidth}
		\centering
		\includegraphics[width=3.4cm,height=2.1cm]{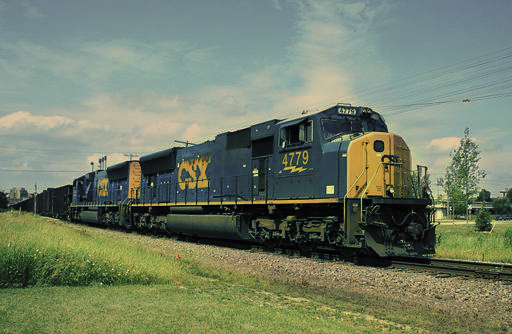}\\
		\scriptsize{HDRNet}
	\end{minipage}
	\begin{minipage}[t]{0.19\textwidth}
		\centering
		\includegraphics[width=3.4cm,height=2.1cm]{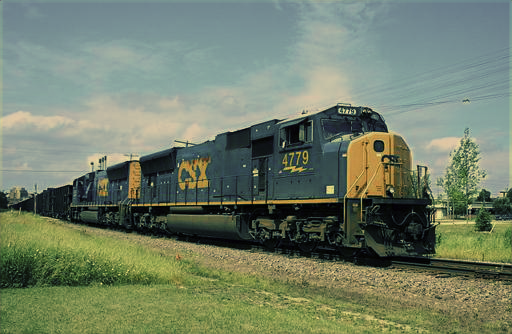}\\
		\scriptsize{CSRNet-L}
	\end{minipage}
	\begin{minipage}[t]{0.19\textwidth}
		\centering
		\includegraphics[width=3.4cm,height=2.1cm]{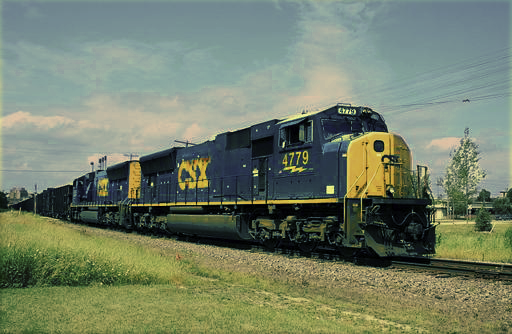}\\
		\scriptsize{GT}
	\end{minipage}

	\begin{minipage}[t]{0.19\textwidth}
		\centering
		\includegraphics[width=3.4cm,height=2.1cm]{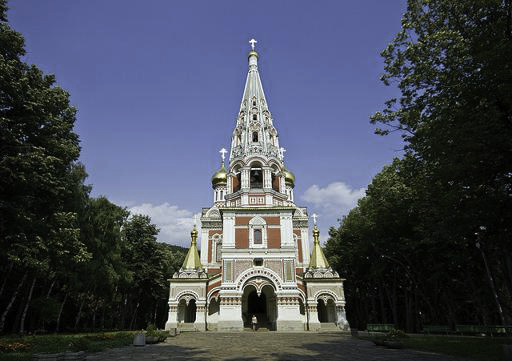}\\
		\scriptsize{(d) input/Watercolor}
	\end{minipage}
	\begin{minipage}[t]{0.19\textwidth}
		\centering
		\includegraphics[width=3.4cm,height=2.1cm]{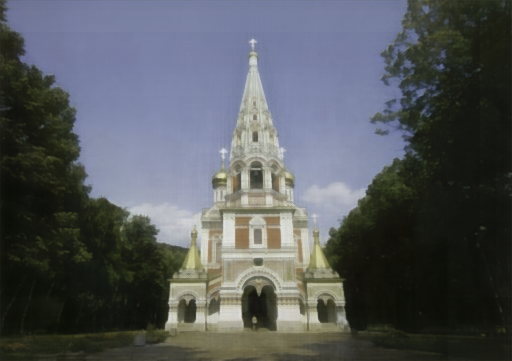}\\
		\scriptsize{CAN}
	\end{minipage}
	\begin{minipage}[t]{0.19\textwidth}
		\centering
		\includegraphics[width=3.4cm,height=2.1cm]{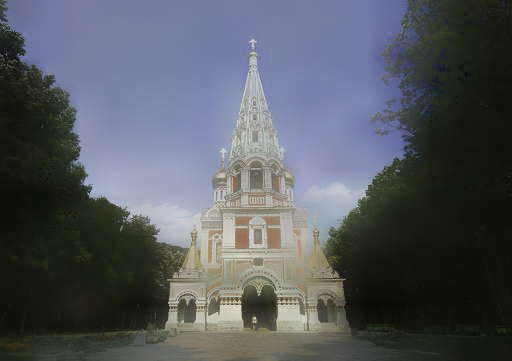}\\
		\scriptsize{HDRNet}
	\end{minipage}
	\begin{minipage}[t]{0.19\textwidth}
		\centering
		\includegraphics[width=3.4cm,height=2.1cm]{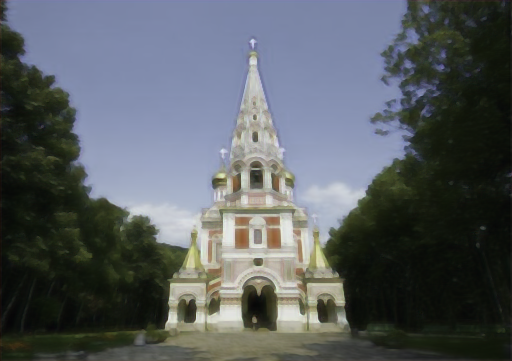}\\
		\scriptsize{CSRNet-L}
	\end{minipage}
	\begin{minipage}[t]{0.19\textwidth}
		\centering
		\includegraphics[width=3.4cm,height=2.1cm]{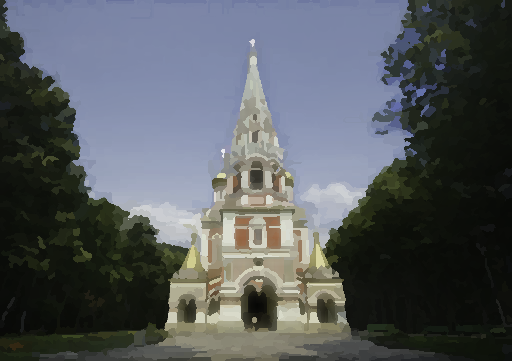}\\
		\scriptsize{GT}
	\end{minipage}

	\caption{Visual comparison on local effect datasets. Row (a): Local Laplacian Filter. Row (b): Foreground Pop-Out effect. Row (c): Local Xpro effect. Row (d): Watercolor effect. The proposed CSRNet-L successfully achieves local adjustments and obtains competitive quantitative and qualitative performance on several local effect datasets. Please zoom in for best view.}
	\label{fig:stylistic}
\end{figure*}

\subsection{Experimental Results}
In this section, we first compare CSRNet-L with a deeper CSRNet version, showing the effectiveness of extending CSRNet from global adjustment to local enhancement. Then we compare with other local enhancement methods, including Pix2Pix and HDRNet. The experimental results reveal that CSRNet-L can achieve comparable performance on several local effect datasets yet with much fewer parameters.
\subsubsection{Extending CSRNet: from global to local}
We demonstrate the effectiveness of extending CSRNet, in terms of enlarging the filter size and adopting spatial feature modulation (SFM). As mentioned above, $1 \times 1$ filters are not able to achieve local adjustment, thus enlarging the filter size in the base network is necessary. We expand the filter size from $1 \times 1$ to $3 \times 3$, so that the base network can learn local patterns and perform local operations. Since enlarging the filter size brings more parameters, we deepen the original CSRNet to 7 layers for fair comparison. Figure \ref{fig:csrnet_local} shows that CSRNet-L can successfully achieve local effects, while CSRNet with $1 \times 1$ filters only performs global adjustment but fails to realize local enhancement. {Quantitatively, for LLF, the global CSRNet only achieves 21.38dB in PSNR and 0.769 in SSIM, while the local CSRNet-L yields 25.03dB and 0.904, respectively.}

\subsubsection{Comparison with other methods}
We compare our results with HDRNet \cite{hdrnet}, CAN \cite{can} and Pix2Pix \cite{pix2pix}. CAN is a learning-based method for  approximating image processing operators. Besides, Pix2Pix is a well-known image-to-image translation framework and HDRNet is a state-of-the-art image enhancement method. We use them as baselines for comparison. For fairness, we retrain their models on each local effect dataset. The PSNR values between input images and ground truth images are shown in Table \ref{table:local}. It numerically reflects the gap between the inputs and the stylized outputs.

Pix2Pix includes over 11 million parameters but obtains the worst performance. The images produced by Pix2Pix usually contains artifacts and incorrect color tones. CSRNet-L only contains 72k parameters (about 1/6 of HDRNet) but reaches 27.13dB on ``Foreground Pop-out'' dataset, which transcends CAN, HDRNet and Pix2Pix by 0.12dB, 0.40dB, and 3.01dB, respectively. Although CAN obtains better PSNR values on LLF dataset, CSRNet-L reaches the best SSIM values. Also, CSRNet-L surpasses CAN in terms of SSIM on ``Xpro'' effect. On the ``Foreground Pop-out'' and ``Watercolor'' datasets, CSRNet-L achieves the best quantitative performance both in PSNR and SSIM.

Visual comparisons are shown in Figure \ref{fig:stylistic}. It can be observed that CSRNet-L successfully implements the effect of local Laplacian filter, which sharpens the input image and enhances the image details. In contrast, HDRNet cannot reproduce this effect well (see the textures in  Figure \ref{fig:stylistic}(a)). As for the “Foreground Pop-out” dataset, CAN and HDRNet fail to accurately highlight the foreground object, as displayed in the second row of Figure \ref{fig:stylistic}. HDRNet tends to confuse the colors in local regions and mix up the foreground and the background. Similarly, CAN and HDRNet cannot produce the watercolor effect well, since the outputs still retain the style of realistic photos. In conclusion, by extending CSRNet, our method can easily achieve local effects and obtain comparable results over other methods with much fewer parameters. This greatly shows the superiority and effectiveness of the proposed framework. More visual comparisons can be found in the supplementary materials.

\subsubsection{Effect of Training strategy}
For local adjustment, we find that the training strategy plays an important role in the final performance. To optimize CSRNet-L, we first train the base network, then add the condition network and train them jointly. Table \ref{table:training_strategy} shows that the two-stage training strategy can obtain better performance than training from scratch. However, if we train the base network and condition network together from scratch, the performance is sometimes even worse than the sole base network. Compared with training from scratch, CSRNet-L with finetuning strategy improves 0.15dB, 0.51dB, 0.36dB, 0.27dB on four local effect datasets, respectively. Besides, it can also improve the stability during training.

\begin{table}[ht]
	\begin{center}
		\caption{Effectiveness of SFM and training strategy.}
		\label{table:training_strategy}
		\small
		\renewcommand{\arraystretch}{1.4}
		\setlength{\tabcolsep}{1.2pt}
		\begin{tabular}{cc|cccc}
			\hline
			Settings & Strategy& LLF & Foreground & Xpro & Watercolor \\
			\hline
			Only Base $3 \times 3$& scratch &22.26 & 26.94 & 29.38 & 24.19 \\
			Base $3 \times 3$ + GFM & scratch &23.75 & 26.82 & 28.96 & 24.13 \\
			Base $3 \times 3$ + SFM & scratch &24.88 & 26.62 & 29.12 & 24.08 \\
			Base $3 \times 3$ + GFM & finetune &23.81 & 26.99 & 29.37 & 24.18 \\
			Base $3 \times 3$ + SFM & finetune &\textbf{25.03} & \textbf{27.13} & \textbf{29.48} & \textbf{24.35} \\
			\hline
			
		\end{tabular}
	\end{center}
\end{table}

\subsubsection{Effectiveness of conditional modulation}
To validate the effectiveness of SFM, we conduct experiments on three different settings: (1) Only base network without modulation. (2) Base network with GFM. (3) Base network with SFM. The results are summarized in Table \ref{table:training_strategy}. When equipped with GFM, the PSNR rises from 22.26dB to 23.81dB on LLF dataset, but on other datasets, the performance does not get better (compare the second and the fifth rows in the table). This explains that GFM has little effect for local enhancement.
However, after we adopt SFM, the performance is greatly improved from a sole base network, with the improvements of 2.77dB, 0.19dB, 0.1dB, 0.16dB on the four local effect datasets, respectively. This shows that SFM plays a crucial role in local adjustment.

\begin{figure}[htbp]
	\centering
	\includegraphics[width=0.9\linewidth]{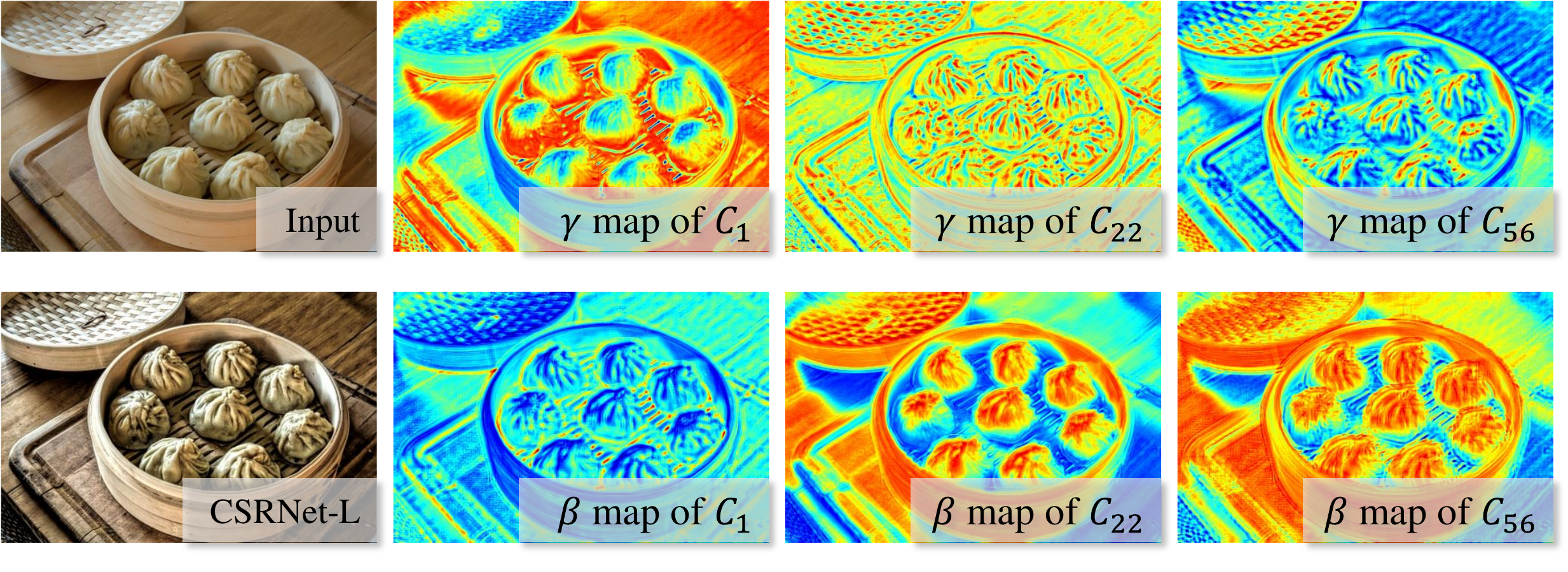}
	\caption{{The modulation parameters $\gamma$ and $\beta$ of CSRNet-L on ``Fast Local Laplacian Filter'' dataset. The parameters contain local spatial information. $C_i$ denotes the $i$-th channel of the first modulation layer.}}
	\label{fig:vis_fea_LLF}
\end{figure}

\begin{figure}[htbp]
	\centering
	\includegraphics[width=0.9\linewidth]{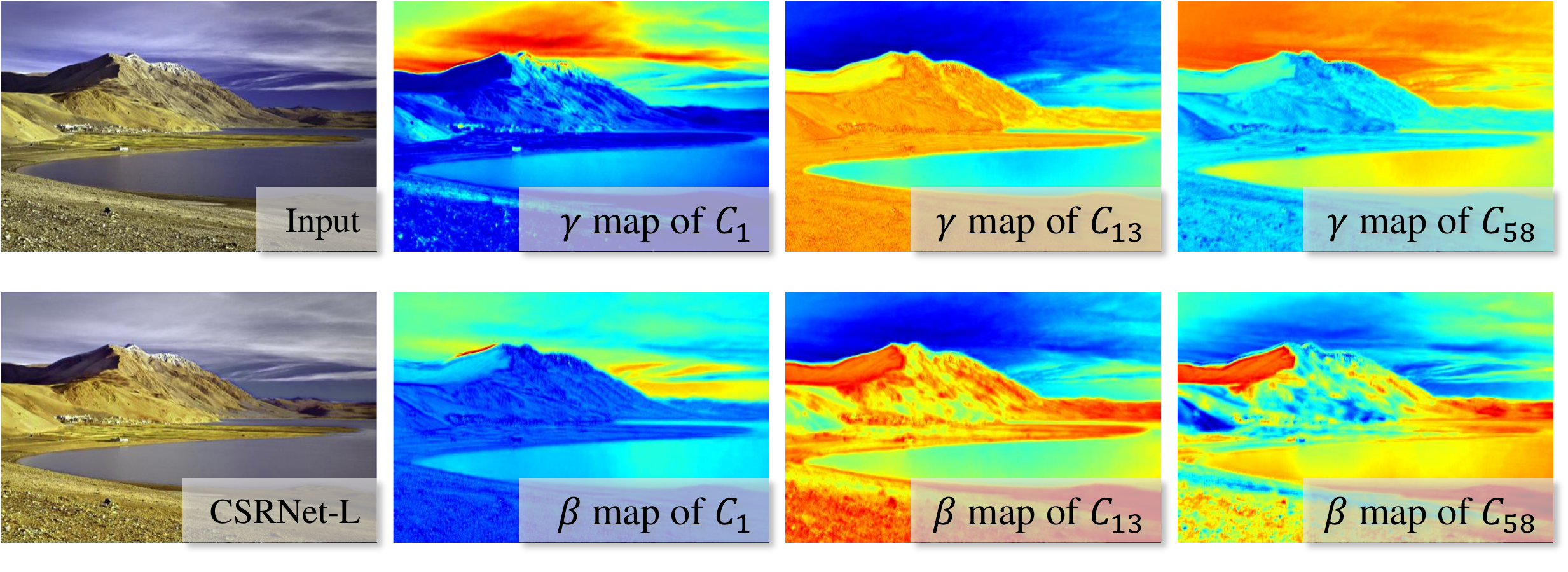}
	\caption{{The modulation parameters $\gamma$ and $\beta$ of CSRNet-L on ``Foreground Pop-Out'' dataset. The parameters distinguish different parts of the image, especially the foreground the background objects.}}
	\label{fig:vis_fea_Foreground}
\end{figure}

{
\subsection{Visualization of Spatial Feature Modulation}
To better understand the behavior of the spatial feature modulation, we visualize the modulation parameters in heatmaps. The modulation parameters of ``Fast Local Laplacian Filter'' (LLF) and ``Foreground Pop-Out'' are shown in Figure \ref{fig:vis_fea_LLF} and Figure \ref{fig:vis_fea_Foreground}, respectively. For LLF effect, the modulation parameters contain obvious spatial information, which well embodies the image corners, edges and textures. Different channels focus on different areas and have different intensities. This corresponds to the operations of local Laplacian filter, which is an edge-aware operator aiming to enhance the image details. For Foreground Pop-Out effect, the modulation parameters reflect more distinct spatial information, which clearly distinguishes different parts of the image, especially the foreground object and the background. For example, in Figure \ref{fig:vis_fea_Foreground}, the $\gamma$ map of $C_1$ mainly concentrates on the clouds, the $\gamma$ map of $C_{13}$ highlights the foreground mountains, and the $\gamma$ map of $C_{58}$ has higher response to the background sky and lake. This suggests that the learned modulation parameters successfully help the network achieve specific local effets.
}

\section{Conclusion}
In this work, we present an efficient image retouching network with extremely fewer parameters. Our key idea is to mimic the sequential processing procedure and implicitly model the editing operations in an end-to-end trainable network. The proposed CSRNet (Conditional Sequential Retouching Network) consists of a base network and a condition network. The base network acts like an MLP for individual pixels, while the condition network extracts global features to generate a condition vector. Then, the condition vector is transformed to modulate the intermediate features of the base network by global feature modulation (GFM). Extensive experiments show that our method achieves state-of-the-art performance on the benchmark MIT-Adobe FiveK dataset quantitively and qualitatively. In addition, besides achieving global tonal adjustment, the proposed framework can be extended to learn local effects as well. By expanding the base network and introducing spatial feature modulation (SFM), our extended method, named CSRNet-L, can successfully achieve local effect adjustment and attain comparable performance against several existing methods.

\section*{Acknowledgments}
This work is partially supported by National Natural Science Foundation of China (61906184), the Joint Lab of CAS-HK, and the Shanghai Committee of Science and Technology, China (Grant No. 20DZ1100800 and 21DZ1100100).


%

\bibliographystyle{IEEEtran}
\bibliography{bio}

\begin{IEEEbiography}[{\includegraphics[width=1in,height=1.25in,clip,keepaspectratio]{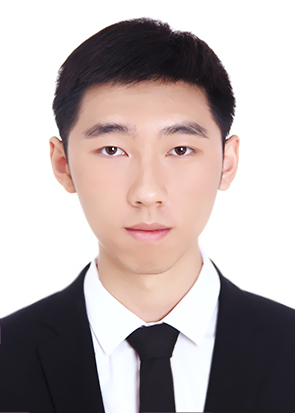}}]{Yihao Liu}
	received the B.S. degree from University of Chinese Academy of Sciences, Beijing, in 2018. He is now working towards the Ph.D. degree in Multimedia Laboratory, Shenzhen Institute of Advanced Technology, Chinese Academy of Sciences. He is supervised by Prof. Yu Qiao and Prof. Chao Dong. His research interests include computer vision and image/video enhancement.
\end{IEEEbiography}

\begin{IEEEbiography}[{\includegraphics[width=1in,height=1.25in,clip,keepaspectratio]{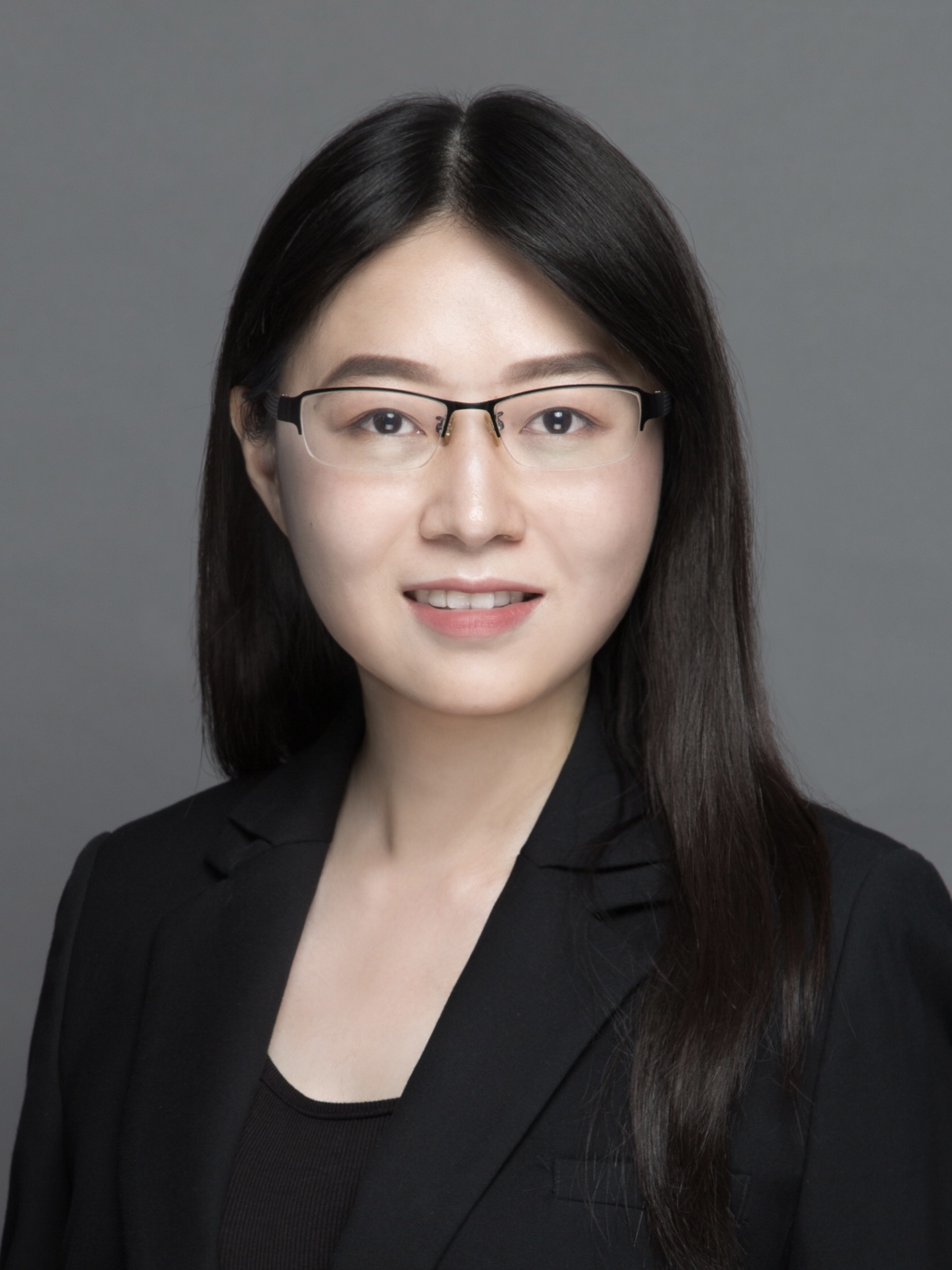}}]{Jingwen He}
	received the B.Eng. degree in computer science and technology from Sichuan University, China, in 2016, and the M.Phil. degree in electronic and information engineering from the University of Sydney, Australia, in 2019. She is currently a Research Intern with the Institute of Advanced Computing and Digital Engineering, Shenzhen Institutes of Advanced Technology, Chinese Academy of Sciences, China. Her current research interests include deep learning and computer vision.
	
\end{IEEEbiography}

\begin{IEEEbiography}[{\includegraphics[width=1in,height=1.25in,clip,keepaspectratio]{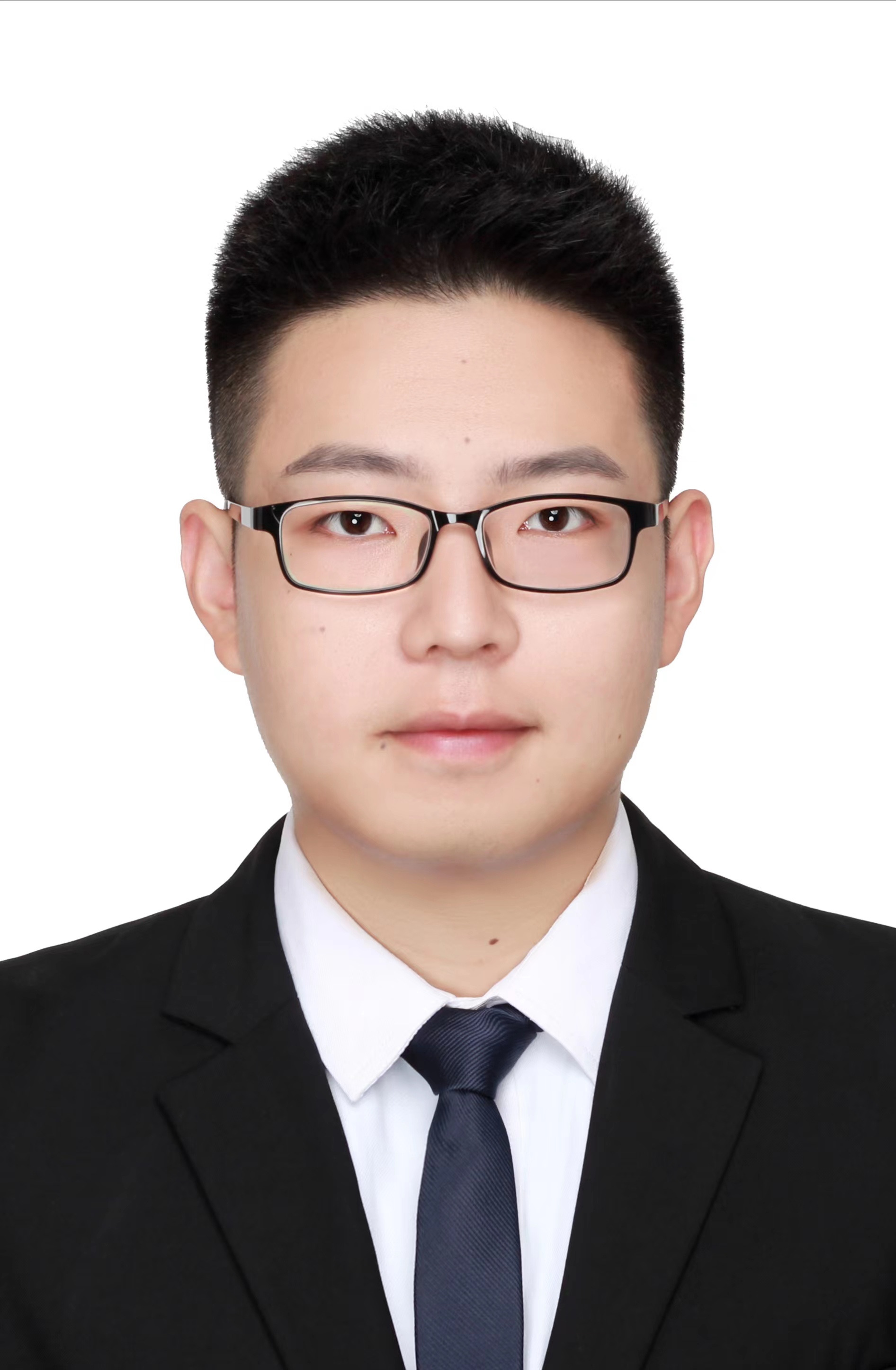}}]{Xiangyu Chen}
	received his B.E. degree and M.E. degree from Northwestern Polytechnical University, Xi'an, China, in 2017 and 2020, respectively. Currently he is a joint PhD. Student in University of Macau and Shenzhen Institute of Advanced Technology, Chinese Academy of Sciences. His research interests include image/video restoration and enhancement. 
	
\end{IEEEbiography}

\begin{IEEEbiography}[{\includegraphics[width=1in,height=1.25in,clip,keepaspectratio]{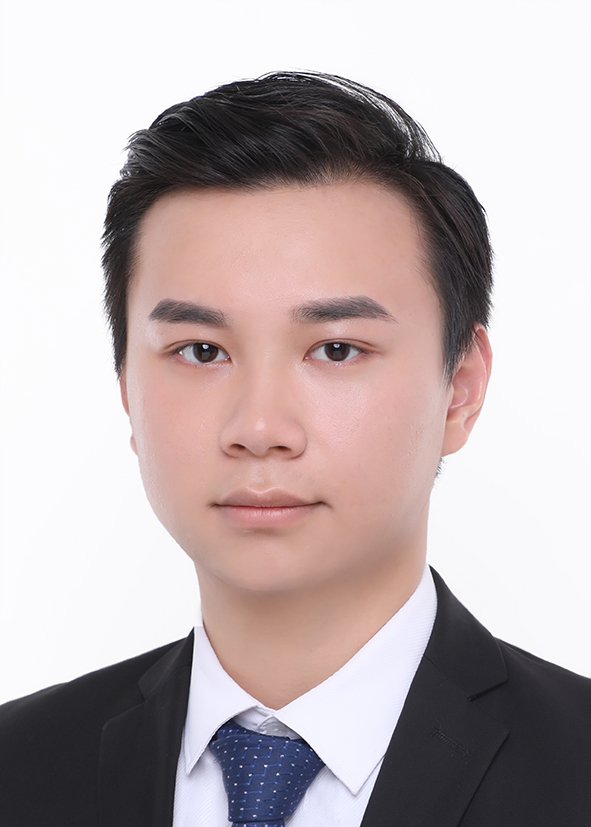}}]{Zhengwen Zhang}
	received the M.S. degree from Beijing Institute of Technology, Beijing, in 2019. He now serves as a research assistant in Multimedia Laboratory, Shenzhen institute of Advanced Technology, Chinese Academy of Sciences. He is supervised by Prof. Yu Qiao and Prof. Chao Dong. His research interests include computer vision, image/video enhancement and weakly supervised learning.
	
\end{IEEEbiography}

\begin{IEEEbiography}[{\includegraphics[width=1in,height=1.25in,clip,keepaspectratio]{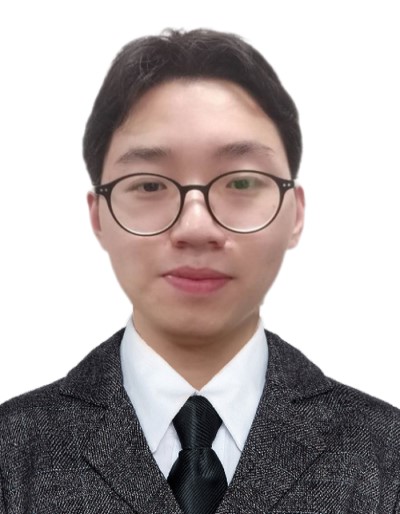}}]{Hengyuan Zhao}
	received the B.S. degree from Nanjing University of Posts and Telecommunications, Nanjing, in 2020. He worked as a research intern in Multimedia Laboratory, ShenZhen Institute of Advanced Technology, Chinese Academy of Sciences. He was supervised by Prof. Yu Qiao and Prof. Chao Dong. His research interests include computer vision and image/video processing and generation.
\end{IEEEbiography}

\begin{IEEEbiography}[{\includegraphics[width=1in,height=1.25in,clip,keepaspectratio]{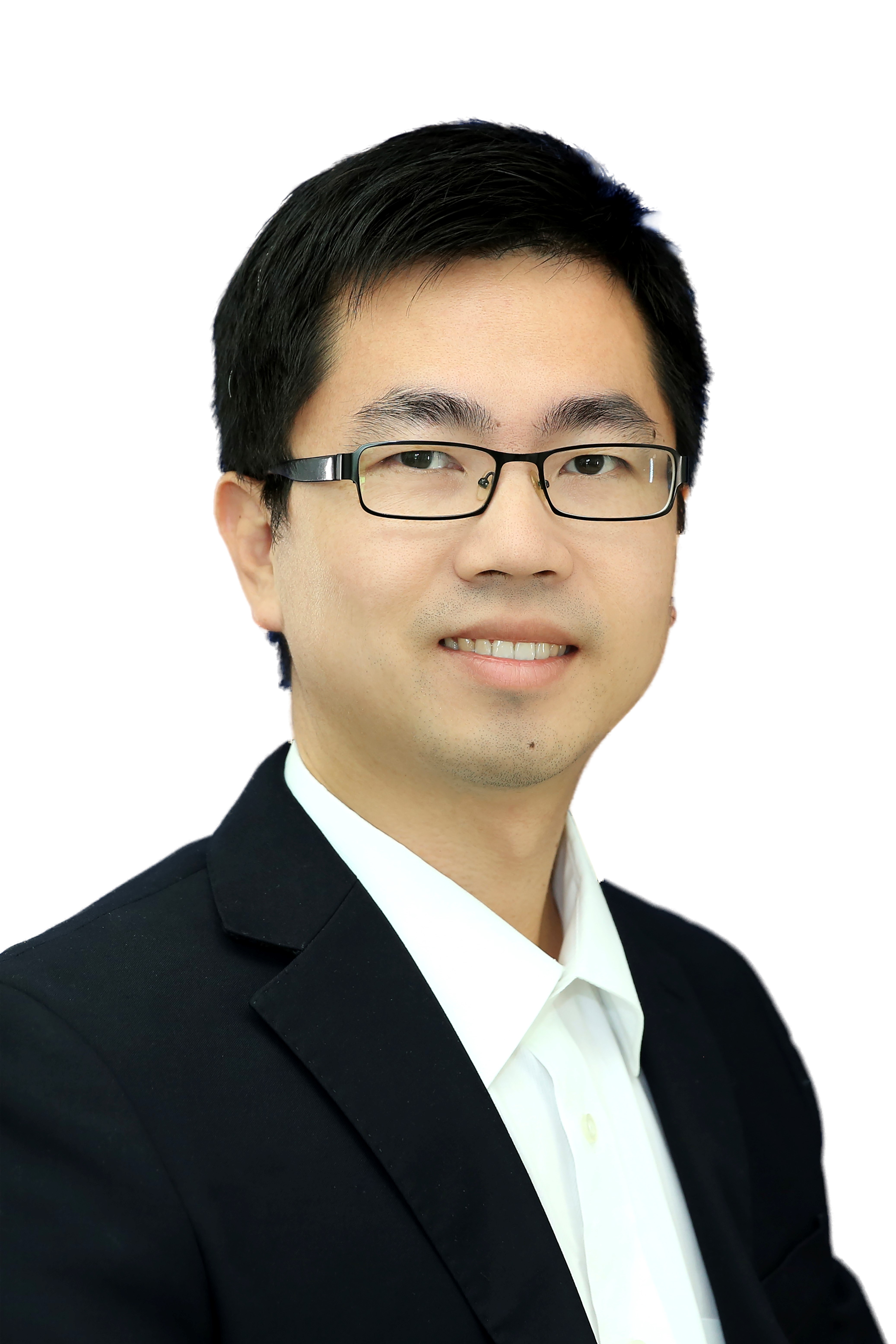}}]{Chao Dong}
	is currently an associate professor in Shenzhen Institute of Advanced Technology, Chinese Academy of Science. He received his Ph.D. degree from The Chinese University of Hong Kong in 2016. In 2014, he first introduced deep learning method -- SRCNN into the super-resolution field. This seminal work was chosen as one of the top ten ``Most Popular Articles'' of TPAMI in 2016. His team has won several championships in international challenges –- NTIRE2018, PIRM2018, NTIRE2019, NTIRE2020, AIM2020 and NTIRE2022. He worked in SenseTime from 2016 to 2018, as the team leader of Super-Resolution Group. In 2021, he was chosen as one of the World’s Top $2\%$ Scientists. In 2022, he was recognized as the AI 2000 Most Influential Scholar Honorable Mention in computer vision. His current research interest focuses on low-level vision problems, such as image/video super-resolution, denoising and enhancement. 
\end{IEEEbiography}

\begin{IEEEbiography}[{\includegraphics[width=1in,height=1.25in,clip,keepaspectratio]{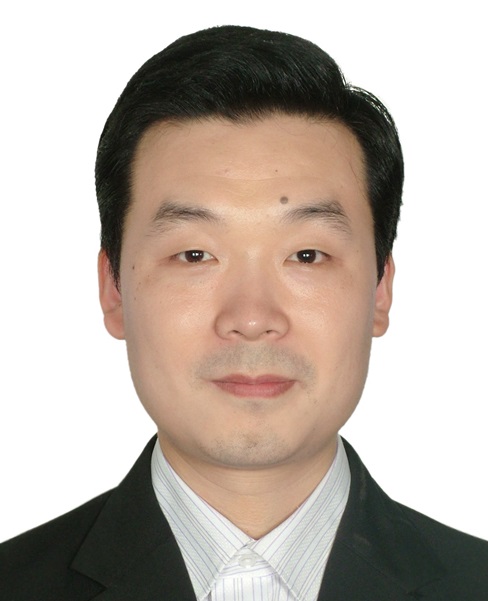}}]{Yu Qiao}
	(Senior Member, IEEE) is currently a Professor with the Shenzhen Institutes of Advanced Technology (SIAT), Chinese Academy of Science, and the Director of the Institute of Advanced Computing and Digital Engineering. He has published more than 180 articles in international journals and conferences, including T-PAMI, IJCV, T-IP, T-SP, CVPR, and ICCV. His research interests include computer vision, deep learning, and bioinformation. He received the First Prize of the Guangdong Technological Invention Award, and the Jiaxi Lv Young Researcher Award from the Chinese Academy of Sciences. His group achieved the first runner-up at the ImageNet Large Scale Visual Recognition Challenge 2015 in scene recognition, and the Winner at the ActivityNet Large Scale Activity Recognition Challenge 2016 in video classification. He has served as the Program Chair of the IEEE ICIST 2014.
	
\end{IEEEbiography}

\vfill

\end{document}